\documentclass[preprint,10pt,authoryear,hidelinks,nonatbib]{elsarticle}
      \usepackage{hyperref}
\usepackage[utf8]{inputenc}
            
          \usepackage[normalem]{ulem}
            \usepackage{float}
           
             \usepackage{subfigure}
             \usepackage{booktabs} 
              \usepackage{graphicx}
             
                    \usepackage{textcomp}
            \usepackage[normalem]{ulem} 
            \usepackage{soul} 
                    \usepackage{epsfig}
                    \usepackage{subfigure}
                    \usepackage{amssymb}
                    \usepackage{amsmath}
                    \usepackage{mathabx}

                    \usepackage{epstopdf}
                    \usepackage{threeparttable}
                    \usepackage{color,soul}
                    \usepackage{xcolor} 
                    \usepackage{amssymb}
                    \usepackage{pifont}
                    \usepackage{enumitem}
                    \usepackage{chngcntr}
                     \definecolor{morado1}{HTML}{37025A}
                      \definecolor{morado2}{HTML}{683E7F}
                       \definecolor{morado3}{HTML}{9779A4}
                        \definecolor{morado4}{HTML}{C9B5CA}
                        \definecolor{morado5}{HTML}{F5EBEB}
                        
                          \definecolor{violeta1}{HTML}{37025A}
                        \definecolor{violeta2}{HTML}{683E7F}
                        \definecolor{violeta3}{HTML}{9779A4}
                        \definecolor{violeta4}{HTML}{C9B5CA}
            
                    \definecolor{gris2}{rgb}{0.92,0.92,0.92}
                    \definecolor{gris1}{rgb}{0.8,0.8,0.8}
                    \definecolor{azul8}{rgb}{0.2039    0.7137    0.6039}
                    
                    \definecolor{azul7}{rgb}{0.0588    0.4431    0.8706}
                    
                    \definecolor{azul1}{HTML}{08306B}
                    \definecolor{azul2}{HTML}{2171B5}
                    
                    \definecolor{azul3}{HTML}{4292C6}
                    
                    \definecolor{azul4}{HTML}{9ECAE1}
                    
                    \definecolor{azul5}{HTML}{DEEBF7}
                    
                    \definecolor{azul6}{HTML}{F7FBFF}

                    \definecolor{naranja}{rgb}{0.639,0.360,0}
                    \definecolor{verde}{rgb}{0.7490,0.7373,0.3725}
                    \definecolor{verde2}{rgb}{0.2314,0.7098,0.6118}
                    \definecolor{verde3}{rgb}{0.5216,0.7294,0.5098}

                    \definecolor{amarillo}{rgb}{0.9686,0.9804,0.0510}
                    
                    \definecolor{marron}{rgb}{0.8314    0.7294    0.3569}
                    \definecolor{marron2}{rgb}{    0.8549    0.7255    0.3059}

            \journal{Expert Systems with Applications}

            \usepackage{xcolor}
            \pagecolor{white}
            
\begin{document}

            \begin{frontmatter}
            
            \title{{General} Machine Learning Models for {Interpreting and} Predicting Efficiency Degradation in Organic Solar Cells}
            
            
            
            %
            
            \author[label1,label2]{David Valiente\corref{cor1}}
            \ead{dvaliente@umh.es}
            \affiliation[label1]{organization={University Institute for Engineering Research, Miguel Hernandez University},
                        addressline={Avenida de la Universidad, s/n}, 
                        city={Elche},
                        postcode={03202}, 
                        country={Spain}}
                        
             \author[label2]{Fernando Rodr\'iguez-Mas}
            \ead{fernando.rodriguezm@umh.es}
             \affiliation[label2]{organization={Communications Engineering Department, Miguel Hernandez University},
                        addressline={Avenida de la Universidad, s/n}, 
                        city={Elche},
                        postcode={03202}, 
                        country={Spain}}
                        
             \author[label3]{Juan V. Alegre-Requena\corref{cor1}}
             \ead{jv.alegre@csic.es}
             \affiliation[label3]{organization={Department of Inorganic Chemistry, Institute of Chemical Synthesis and Homogeneous Catalysis (ISQCH), CSIC, University of Zaragoza},
                        addressline={Pedro Cerbuna 12}, 
                        city={Zaragoza},
                        postcode={50009}, 
                        country={Spain}}
            
             \author[label3]{David Dalmau}
             \ead{ddalmau@unizar.es}
             
              \author[label1]{Mar\'ia Flores}
             \ead{m.flores@umh.es}
            
             \author[label1]{Juan C. Ferrer}
             \ead{jc.ferrer@umh.es}
            \cortext[cor1]{Corresponding author}

            \begin{abstract}
\small
Within the range of renewable energy sources, photovoltaic (PV) plays a crucial role in addressing the exponential growth in global demand during this decade. Organic solar cells (OSCs) emerge as a promising alternative to silicon-based PVs due to their low cost, lightweight, and sustainable production. Despite achieving power conversion efficiencies (PCEs) exceeding values of 20\%, OSCs still face challenges in stability and efficiency. In this context, recent advances in manufacturing, artificial intelligence (AI)-driven and machine learning (ML) models offer opportunities to optimize and screen OSCs for greater sustainability and commercial viability, thus potentially reducing costs while ensuring stable and long term performance.~This work presents a set of optimal ML models to represent the temporal degradation suffered by the PCE of polymeric OSCs with a multilayer structure ITO/PEDOT:PSS/P3HT:PCBM/Al. To that aim, we generated a database with {166} entries {with measurements of 5 OSCs}, which includes up to 7 variables regarding both the manufacturing process and environmental conditions for more than 180 days. Then, we relied on a software framework that brings together a conglomeration of automated ML protocols that execute sequentially against our database by simply command-line interface.~This easily permits hyper-optimizing the ML models through exhaustive benchmarking so that optimal models are obtained. The accuracy {for predicting PCE over time} reaches values of the coefficient determination (R$^2$) widely exceeding 0.90, whereas the root mean squared error (RMSE), sum of squared error (SSE), and mean absolute error (MAE) {are significantly low}. Additionally, we {assessed the predictive ability of the} models {using an unseen OSC as an external set}. For comparative purposes, classical Bayesian regression fitting based on non-linear least squares (LS) are also presented, which only perform sufficiently for univariate cases of single OSCs.
            \end{abstract}
            
            %
            \end{frontmatter}

            \section{Introduction}\label{sec:intro}
            
Electricity consumption has evolved exponentially over the last three decades, increasing from a total estimate of 15277 to 29479 TWh between 2000 and 2023, respectively~\cite{Rit2023}. One of the immediate consequences has been reflected in the price of this energy. Simultaneously, in the technological realm, there has been a notable shift in electricity generation towards renewable sources. Particularly, the maturity of conventional photovoltaic (PV) cells and the emergence of new organic and hybrid materials have made it possible to get significant improvements in PV generation and its power conversion efficiency (PCE)~\cite{NREL}. However, {even though} the global consumption of PV energy {has} increase{d} from 1.08  to {1629.9} TWh between 2000 and {2023}~\cite{Rit2023}, its generation still falls short of meeting current demands. Hence, there are many challenges and open lines for improving the generation and efficiency of PV cells.

In this context, organic solar cells (OSCs) have emerged as a promising alternative to silicon-based solar cells since, amongst other benefits, their production involves low-temperature manufacturing methods~\cite{Seo2015} and reduced carbon footprint. Their appeal also lies in the ease of processing, low cost, flexibility, and lightweight~\cite{Yeh2013}. Moreover, the evolution of their PCEs has been much superior to classical technologies and it has been on par with other considered emerging technologies, presenting efficiencies that exceed values of 20\%~\cite{veintePCE}. {Even though} the commercialization of OSCs still faces various challenges, including stability and efficiency, some cost studies~\cite{PVCost1,PVCost2,FER_Heliyon} indicate that once these obstacles are overcome, OSCs can be manufactured at a cost lower than {one} dollar per peak Watt~\cite{Bra2005}.

The integration of machine learning (ML) before transferring to industrial implementation is being extensively tested these days. For instance, within the field of cheminformatics, it is of paramount importance in the design of new materials. Numerous studies have demonstrated the potential of deep learning algorithms to analyze complex chemical data, thereby accelerating the process of discovering, for example, drugs~\cite{Dalmau2024}, as well as identifying promising compounds with improved properties{~\cite{Lus2013,Sanosa2024}}. Some early studies~\cite{Han2013,Rup2012} laid the foundations for the application of ML models in predicting molecular properties, highlighting their great strengths. More recently, the deployment of artificial intelligence (AI) techniques with high-performance prediction methods has enabled the emergence of software platforms capable of examining extensive chemical databases to identify potential materials with significant properties~\cite{Unt2014,Wall2015}. {Furthermore, researchers} moved forward from general chemical structures to the inclusion of PV features{~\cite{Joh2011}, predicting} the electrical behavior of PV devices in terms of parameters such as voltage, current, conversion efficiency, short-circuit current, open-circuit voltage, etc.~\cite{Eib2021,Ser2016,Tol2019,Lop2016}. {Currently,} there is {a wide array} of ML and deep learning-based engines~\cite{Moo2022,Eib2021,Mal2021,Seo2019,Sun2019} {designed} to optimize different target features of optoelectronic devices~\cite{Miy2021,Mah2022a,Mah2022b}.

{Other authors concentrated on AI-supported models to propose long lifecycle materials for sustainable energy generation ~\cite{Vel2023,Ju2018,Mah2023}. Some studies have used ML for predicting optimal installation of energy harvesting systems ~\cite{HeliyonReview1,PV_forecast_EAAI,HeliyonReview2} whereas others have focused on predictors of energy generation~\cite{HeliyonPV,ML_predictUSA,PV_forecast_EAAI}. Likewise, fault detection of PV systems has been also anticipated thanks to automated models ~\cite{DefectVisual,DefectEAAI}.} Others delved into the manufacturing variables of {the} specific {PV} structures~\cite{ML_structure,Power_predict,HeliyonPerov,ML_OSC}. In the same way, there is incipient research on the analysis of stability and durability of these devices~\cite{PerovBig,Lifetime,PEDOT_stability,PerovML}.

In sight of all these developments, this work explores different ML models to encode the PCE performance of polymeric OSCs based on multilayer ITO/PEDOT:PSS/P3HT:PCBM/Al. To that aim, we have used {the} ROBERT {program}~\cite{ROBERT}, {which automates data curation, screening of ML models, assessment of predictive ability, and feature analysis} (for reproducibility details, see~\ref{sec:app3}). Our dataset consists of PCE measurements acquired for 180 days, {of 5 OSC devices}, {resulting in} a database with 166 entries. {Although these devices are encapsulated,} the database contains up to seven descriptors, {some} related to environmental conditions (temperature, hummidity, dew point and pressure) {to inspect possible dependencies on their future performance} { and some others related to their} composition and manufacturing (quantities of: solvent in the HTL layer, i.e., PEDOT:PSS; P3HT, PCBM; and volume ratio of P3HT:PCBM). The accuracy metrics demonstrate the validity of these models to learn and represent the temporal behavior of our OSCs, with determination coefficient up to R$^2$=0.96, and {low} root mean squared error (RMSE), sum of squared error (SSE) and mean absolute error (MAE). Additionally, traditional Bayesian regression models sustained by {non-linear least squares} have been introduced to confirm the advantages of the ML-automated protocols against these classical approaches. While these conventional methodologies~\cite{PVBayes, PVStatistical,PVStatistical2} may yield satisfactory results in univariate regression domains, our approach demonstrates that computational learning modeling offers a much broader and comprehensive solution compared to those based on classical statistical models. {ML models show} increased robustness and reliability for:
\begin{itemize}
\item {Generating multivariate models to capture PCE behavior over time while ensuring the interpretability of variable influences.}
\item Learning from an entire dataset {of multiple OSCs}, not solely from a specific OSC device as in classical regression.
\item Screening new OSC devices not used during the learning phase, thereby predicting its behavior, as an unknown device.
\item {Detecting} optimal variables to fabricate the OSC{s with optimal} PCE and/or its stability over time. 
\end{itemize}

In summary, the key contributions of this research are as follows:
\begin{enumerate}
\item Assessment of optimal ML models that learn the PCE behavior over time for polymer-based OSC devices with ITO/PEDOT:PSS/P3HT:PCBM/Al structure{s}.

\item Comparative benchmarking among different ML models and classical statistical regression approaches.

\item Identification of an optimal ML model capable of predicting the PCE behavior of unseen OSC devices.

\item {Feature} analysis to establish the influence {of variables on} the performance of the OSC devices.

\item Reproducibility and transparency of the obtained ML models using a standardized framework for command line replication.
\end{enumerate}
The rest of the paper is organized as follows: Section~\ref{sec:model} describes the specific OSCs manufactured in our laboratory, their electrical parameters, and the periodic measurements that comprise the database used by the ML methods. Subsequently, these methods are defined, starting from a preliminary scope of regression problems, along with the software framework that enables their benchmarking and, consequently, the extraction of models with optimal hyper-parameters. Then, Section~\ref{sec:results} outlines the experiments and results. Finally, Section~\ref{sec:conclusions} draws conclusions from this work.
            \section{Materials and Methods}\label{sec:model}
            
            \subsection{Organic Solar Cells}\label{sec:OSC}

The OSCs characterized in this work are manufactured through spin-coating technique, which involves the deposition of overlapped polymeric thin films. Polymers are deposited in solution, then a process of rotation followed by heating aids in removing the solvent, hence forming the layer. The structure of the devices was as follows: ITO/PEDOT:PSS/P3HT:PCBM/Al, where each layer refers to:
\begin{itemize}
\item ITO: Indium tin oxide.
\item PEDOT:PSS: poly(3,4-ethylenedioxythiophene) polystyrene sulfonate.
\item P3HT:PCBM: (poli(3-hexiltiofeno-2,5-diil):[6,6]-phenyl-C61-butyric acid methyl ester.
\item Al: Aluminium.
 \end{itemize}
 As a substrate, a glass with a 60 nm thick semi-transparent ITO layer was used. The substrates {were} placed in the spinner with the ITO layer facing upwards so that the thin layers {could} be deposited on it. The first layer was the PEDOT:PSS film. It was deposited at room temperature at a rotation speed of 6000 rpm for 60 seconds. The remaining solvent was removed by heating at 150~$^{\circ}$C for 10 minutes.
Once the first layer was deposited, the active layer composed by the P3HT:PCBM polymer blend was applied. This was done at 300 rpm for 3 minutes. It was then dried at 80~$^{\circ}$C for 1 hour.

The final layer deposited on the devices {was an} aluminium film. Aluminium {was} not deposited using the spin-coating technique but rather by metal evaporation in a vacuum chamber. The equipment consists of a high vacuum chamber, two vacuum pumps, one of which is a high vacuum pump, and a power supply to provide the necessary current to evaporate the aluminium. Once the samples {were} placed inside the chamber, vacuum {was} achieved, reaching pressures of 10$^{-6}$ mbar. When this pressure {was} reached, the aluminium evaporated by Joule effect.
           
\subsection{OSC database}\label{sec:database}
    
 As initially mentioned in Section~\ref{sec:intro}, our dataset consists of PCE measurements of {5 OSC} devices, where up to 7 variables {were} registered for more than 180 days. Electrical parameters (J-V curves) of these OSCs {were} acquired along with climate conditions. The acquisition process comprises a Keithley-2400 equipment that acts as source generator for the voltage sweep as well as recorder of the generated PV current. The J-V curves {were} measured under light conditions (100 mW/cm$^2$, AM 1.5~G, and 25~$^{\circ}$C) generated by a solar simulator Newport xenon arc lamp and an AM 1.5G filter. The electrical characterization {was} completed by determining the characteristic electrical parameters, including the short-circuit current density (J$_{sc}$), the open-circuit voltage (V$_{oc}$), the maximum power point (P$_{mpp}$), the fill factor (FF), and the PCE. Obtaining these parameters allow{ed} to extract the PCE value as follows:
    \begin{equation}
PCE = \frac{P_{mpp}}{P_{inc}}=\frac{J_{sc}V_{oc}FF}{G}
\label{eq:PCE}
\end{equation}
being P$_{inc}$ the incident solar power on the OSCs, which derives from the incident irradiance of the solar simulator, G.

\begin{figure}[h!]
                \centering
                \includegraphics[scale=0.3,clip]{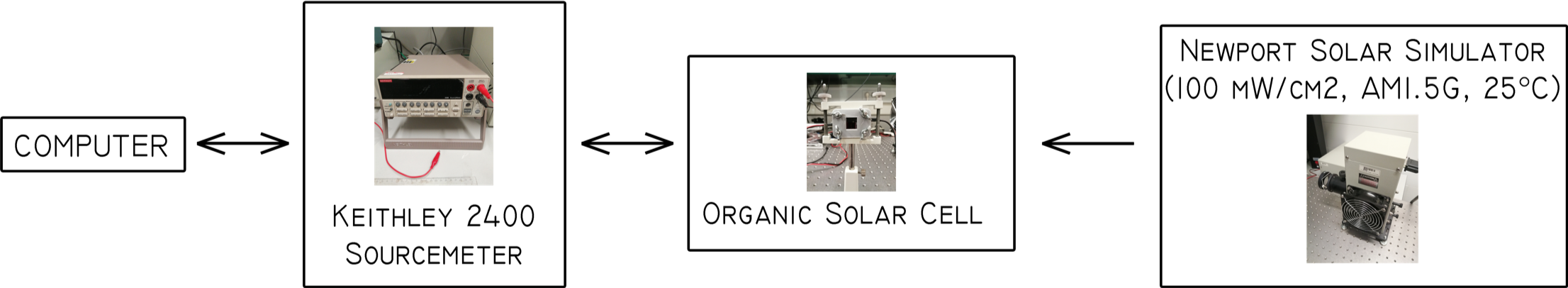}
               \caption{Block diagram of the acquisition system used to acquire the OSC database.}
               \label{fig:aquisition}
            \end{figure}

The block diagram of the equipment for the acquisition system is presented {in} Figure~\ref{fig:aquisition}. This permitted obtaining a database with {166 entries}. The variables and their units are presented {in} Table~\ref{tab:table1}. {Notice that, in order to get further insights on the future performance of the OSCs, environmental conditions at laboratory have also been measured, despite the devices were encapsulated.}

\begin{figure}[h!]
                    \centering
                            \subfigure[]{\includegraphics[clip,width=0.49\columnwidth, height=0.49\columnwidth]{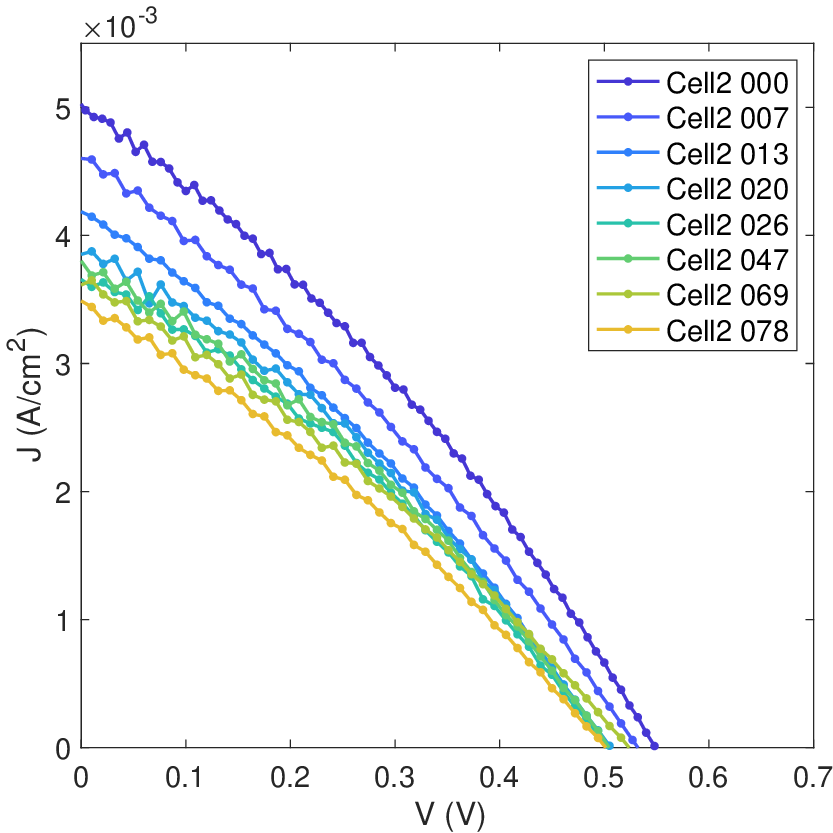}\label{fig:a}}
                    \subfigure[]{\includegraphics[clip,width=0.49\columnwidth, height=0.49\columnwidth]{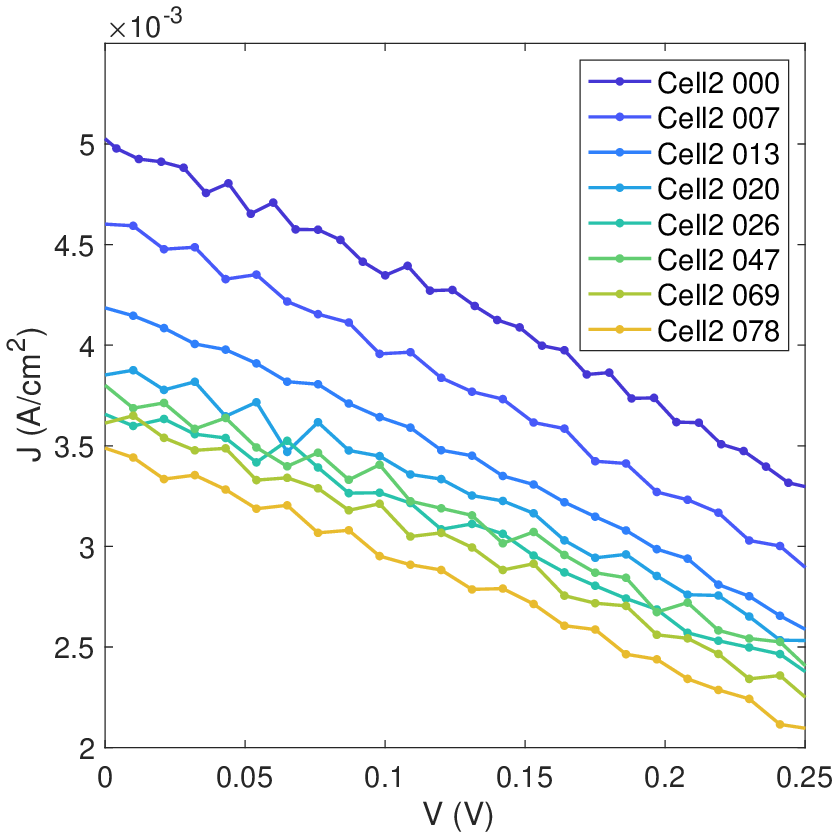}\label{fig:b}}
                    \caption{Evolution of current density (J) vs. voltage (V) over time (days) for the OSC $Cell2$.~(\textbf{a}) Curve J-V with V~$\in$[0-0.7] V.~(\textbf{b}) Same curve J-V with V~$\in$[0-0.25] V.} 
                    \label{fig:IVs}
                    \end{figure}

            \begin{table}[h!]
                    \caption{Detail of the dataset: {Manufacturing variables and environmental conditions at laboratory.}}
                    \label{tab:table1}
                    \centering
                    \footnotesize
                    \begin{tabular}{c|c}
                    \toprule
                    \textbf{Variables}&\textbf{Values}\\
                    \midrule
                    solvent quantity HTL (PEDOT:PSS) [${\mu}$l]	        &[250-1000]\\
		   P3HT [mg]	        &[1-1.2]  \\ 
		   PCBM [mg]	        &[0.8-1] \\ 
                    Volume ratio P3HT:PCBM	        &[1-1.25]\\ 
                    Temperature [$^{\circ}$C]	        &[12-23]\\   
                    Hummidity [\%]	        &[33-88]\\ 
                    Dew point [$^{\circ}$C]	        &[3-19]\\
                    Pressure [hPa]	        &[997-1022]\\
                    Time [days]	        &[0-181]\\
                     
                    \bottomrule
                    \end{tabular}
                    \end{table}
\pagebreak
As an example, Figure~\ref{fig:IVs} presents an OSC {device} contained in the database, for which its current density (J) data have been periodically acquired against voltage (V) over {time}. Specifically, Figure~\ref{fig:a} displays the voltage range on the X-axis from 0 to 0.5 V, while Figure~\ref{fig:b} adjusts the scale to show the range from 0 to 0.25 V, allowing for a clearer observation of the evolution of the J-V curve {when days go by (different colors of the legend)}. The degradation principles of these organic devices suggest that over time, these curves, from which maximum power and energy conversion efficiency are obtained, should consistently {decay}. 

\begin{figure}[t!]
                \centering
                \includegraphics[scale=0.45,clip]{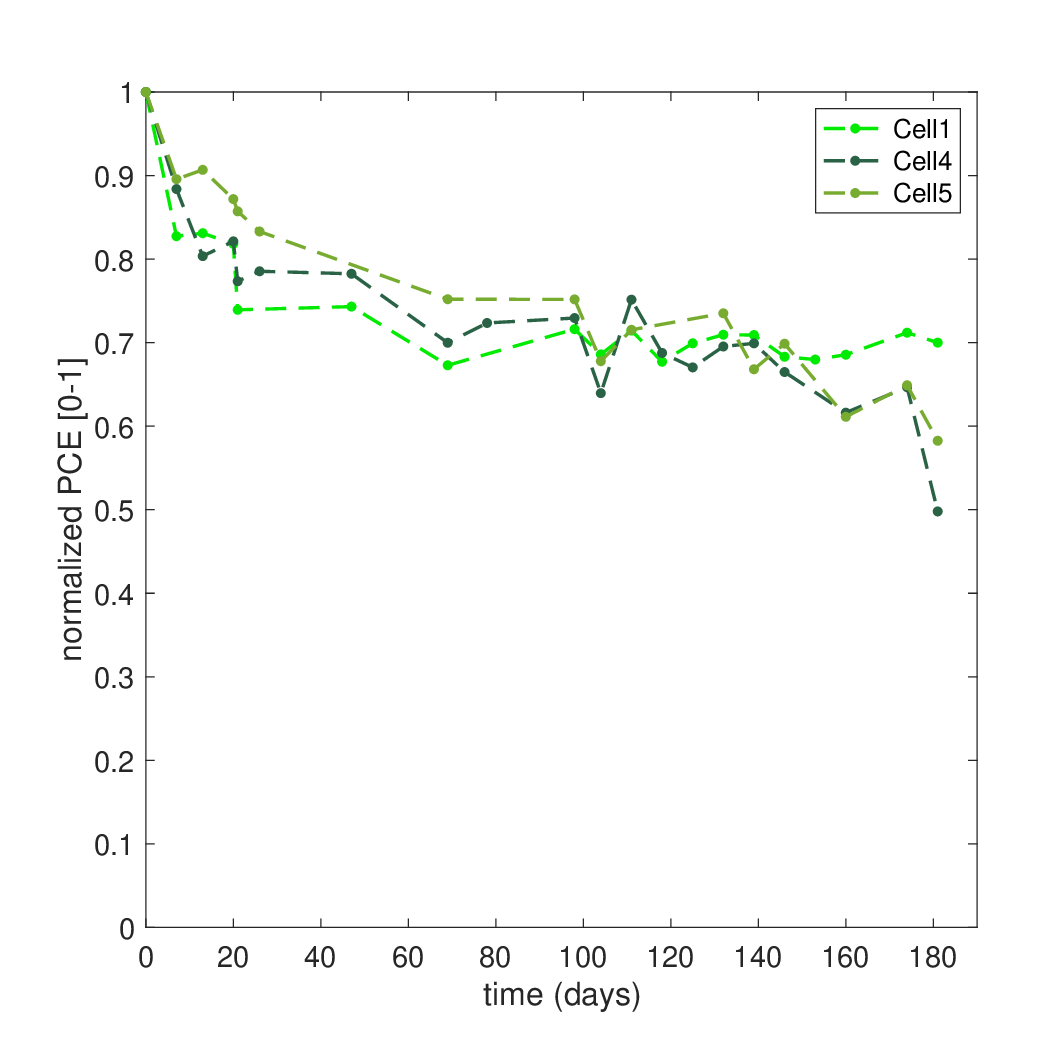}
               \caption{Evolution of the normalized PCE over time for three different OSCs.}
               \label{fig:normPCE}
            \end{figure}

{Likewise}, Figure~\ref{fig:normPCE} compares the temporal evolution of three different OSCs in terms of their normalized PCE values, ranging from 0 to 1, for more than 180 days. {Please note that normalization is used for visualization purposes, as typically used in this field~\cite{Wurfel} when analyzing PCE decay}. Once again, it is confirmed that time leads to the degradation of the device's conversion efficiency. However, it also emerges that differences in the fabrication of the OSCs, as well as the environmental conditions during their measuring, may influence the trend followed by the PCE over time.
\clearpage
              \subsection{Regression Problem {with Classical Approaches}}                 

{T}his study aims to automatically model the degradative behavior exhibited by manufactured OSCs using ML approaches{, which is} a regression problem. {I}n order to establish further comparisons that highlight and reinforce the abilities of {ML}-based protocols, a preliminary step {was} carried out to present regression fitting using traditional {LS}-supported Bayesian methods.

Regression is a method for estimating the relationship between a response or output variable and one or more predictor or input variables. Linear and non-linear regression serve as estimators of values between observed data points. From that starting point, a regression model relates response data to predictor data through one or more coefficients. Then a {parametric} fitting algorithm is needed to calculate some model coefficients from a set of input data. Therefore, {the} parametric {algorithm} estimates deterministic component{s}, whereas random component{s are} typically described as the error. Considering the model as a coefficient function of the independent variable, the error encodes random fluctuations around a Gaussian probability distribution. In many scopes, the goal is to minimize that error, classically addressed by means of {least squares (LS)} fitting approaches, such as: linear least mean squares, weighted least squares, robust least squares, non-linear least squares, etc. 
\begin{equation}
y = f(X, \beta) + \varepsilon
\label{eq:leastsquares}
\end{equation}
where $y$ is the output vector data of n$\times$1, corresponding to the input data in $X$ of n$\times$m, after being applied $f$ as a non-linear function of the coefficient vector $\beta$ (m$\times$1), being $\varepsilon$ the vector of unknown errors of  n$\times$1. Afterwords, the SSE is minimized, understood as the residual sum of squares, given a set of $n$ data values, the residual value of the $i$-{th} value $r_i$ is calculated as:
\begin{equation}
 r_i= y_i-\hat{y}_i
 \end{equation}
 where $y_i$ represents the $i$-th observed value and $\hat{y}_i$ represents the $i$-th estimated value, and accordingly:
\begin{equation}
SSE = \sum_{i=1}^{n} r_i^2 = \sum_{i=1}^{n} (y_i - \hat{y}_i)^2
\label{eq:SSE}
\end{equation}
Subsequently, the algorithm proceeds iteratively calculating the coefficients from an initial seed. Sometimes non-linear models trust on heuristic schemes to calculate initial values. For others models, coefficients randomly initialized in ranges from [0-1]. Then the response value is given as \(\hat{y}=f(X, \beta) \), computed using the jacobian matrix of \( f(X, \beta) \), as the matrix that contains the partial derivatives with respect to the coefficients of \( \beta \). Finally, the adjustment of the coefficients for the next iteration lies on some non-linear least squares algorithms, such as Levenberg-Marquardt, Gradient descent or Gauss-Newton~\cite{LMS}. Whenever the fitting meets the specified convergence criteria, the final solution is assumed as valid.

After observing in Section~\ref{sec:database} the behavior of PCE over time for the various OSCs contained in our database, it {suggests} that the {LS} regression fitting models yielding the best results {are} those with non-linear characterization. Table~\ref{tab:table2} displays the selected models along with their expressions as a function of time, dependent on the adjustment coefficients. It should be noted that the capability of these classical models lies solely in modeling the univariate behavior of the time effect on the PCE values. While it is possible to explore other {LS} regression fitting in the multivariate domain, they only allow for establishing linear relationships, which do not adequately capture the behavior of our devices.
            \begin{table}[th!]
                    \caption{{LS} Bayesian regression fitting models to estimate PCE, denoted as $f(x)$, where $x$ represents time by means of parametric models.}
                    \label{tab:table2}
                    \centering
                    \footnotesize
                    \begin{tabular}{c|c}
                    \toprule
                    \textbf{Parametric model}&\textbf{Coefficient expression of $f(x)$}\\
                    \midrule
                    		$exp1$&$a e^{b x}$\\
                   		$exp2$&$a e^{b x}+c e^{d x}$\\
                   		$gauss1$& $a_1 e^{-[(x-b_1)/c_1]^2}$\\
            		$gauss2$&$a_1 e^{-[(x-b_1)/c_1]^2} + a_2 e^{-[(x-b_2)/c_2]^2}$\\
            		$poly3$&$p_1 x^3 + p_2 x^2 + p_3 x + p_4$\\
               
                    \bottomrule
                    \end{tabular}
                    \end{table}
\subsection{ML framework}\label{sec:ROBERT}

In contrast to the previous classical approaches, ML moves forward to produce non-parametric regression models that adjust more complex behaviors. In this work, we exploit the advantages of a software framework developed under Python, ROBERT~\cite{ROBERT} (see~\ref{sec:app3} for reproducibility details), that facilitates hyper-optimization and benchmarking over well-recognized ML regression models by single command line instruction. This {automated} framework consists of the following modules (for further details please check the {online documentation~\cite{Readthedocs}):}
\begin{itemize}
\item {CURATE (d}ata curation): It processes the {input} dataset in order to filter correlated descriptors, noise, duplicates, as well as to identify and to convert categorical variables into one-hot descriptors.

\item {GENERATE (m}odel selection): It iterates through multiple hyper-optimized models from scikit-learn~\cite{pedregosa11a}{, including Random Forest (RF), Multivariate Linear Model (MVL), Gradient Boosting (GB), Gaussian Process (GP), AdaBoost Regressor (AdaB), MLP Regressor Neural Network (NN), and Voting Regressor (VR). The algorithms are combined with different {training-validation split} sizes, from 60-40\% to 90-10\% using random data splitting. For each combination of algorithm and training size, two models are generated: one with all the descriptors and another with only the most important variables detected by permutation feature importance (PFI) analysis. Among all the possibilities, the program selects two optimal models based on RMSE error.}


\item {PREDICT (external predictions selection): the framework is able to predict new target values. Moreover, it provides feature importances using SHapley Additive exPlanations (SHAP) {and Permutance Feature Importance (PFI) analysis, and outlier detection.}}

\item {VERIFY (assesing predictive ability): It assesses the predictive ability of the models, considering tests such as y-shuffle, y-mean, k-fold cross-validation, and prediction with one-hot encoding.}

\item {REPORT (generation of PDF reports): With the aim to enhance reproducibility and transparency, this module offers a detailed report containing comprehensive information about the ML models utilized and replication instructions through command line executions.}
\end{itemize}
%
            \section{Results}\label{sec:results}

This section introduces the results obtained using both classical {LS} regression fitting {and ML} models to estimate the temporal behavior of our OSCs. The selected error metrics for analysis are briefly presented below:
\begin{itemize}
\item Coefficient of determination R$^2$: It quantifies how well the independent variables explain the variability of the dependent variable. Higher values indicate that the model fits the data well and captures {a} larger proportion of the variability in the dependent variable.
\begin{equation}
R^2 = 1 - \frac{SSE}{SS_{tot}}
\end{equation}
where SSE was defined in (\ref{eq:SSE}) and SS$_{tot}$ is the total sum of squares.
\item RMSE: It provides a measure of the average magnitude of the errors made by the model in its predictions. Minimizing this error is often a goal when training regression models in ML.
\begin{equation}
RMSE = \sqrt{\frac{1}{n} \sum_{i=1}^{n} (y_i - \hat{y}_i)^2}
\end{equation}
where $n$ is the number of samples, $y_i$ is the real observed value and $\hat{y}_i$ is the predicted value.
\item MAE: Another common objective error that measures the average magnitude of the errors between the predicted values and the actual values of the target variable. Unlike RMSE, which penalizes large errors more heavily, MAE treats all errors equally by taking the average of their absolute values.
\end{itemize}
\begin{equation}
MAE = \frac{1}{n} \sum_{i=1}^{n} |y_i - \hat{y}_i|
\end{equation}

            \subsection{Classical {LS} regression fitting results}\label{sec:bay}

\begin{figure}[h!]
                    \centering
                            \subfigure[]{\includegraphics[clip,width=0.49\columnwidth, height=0.30625\columnwidth]{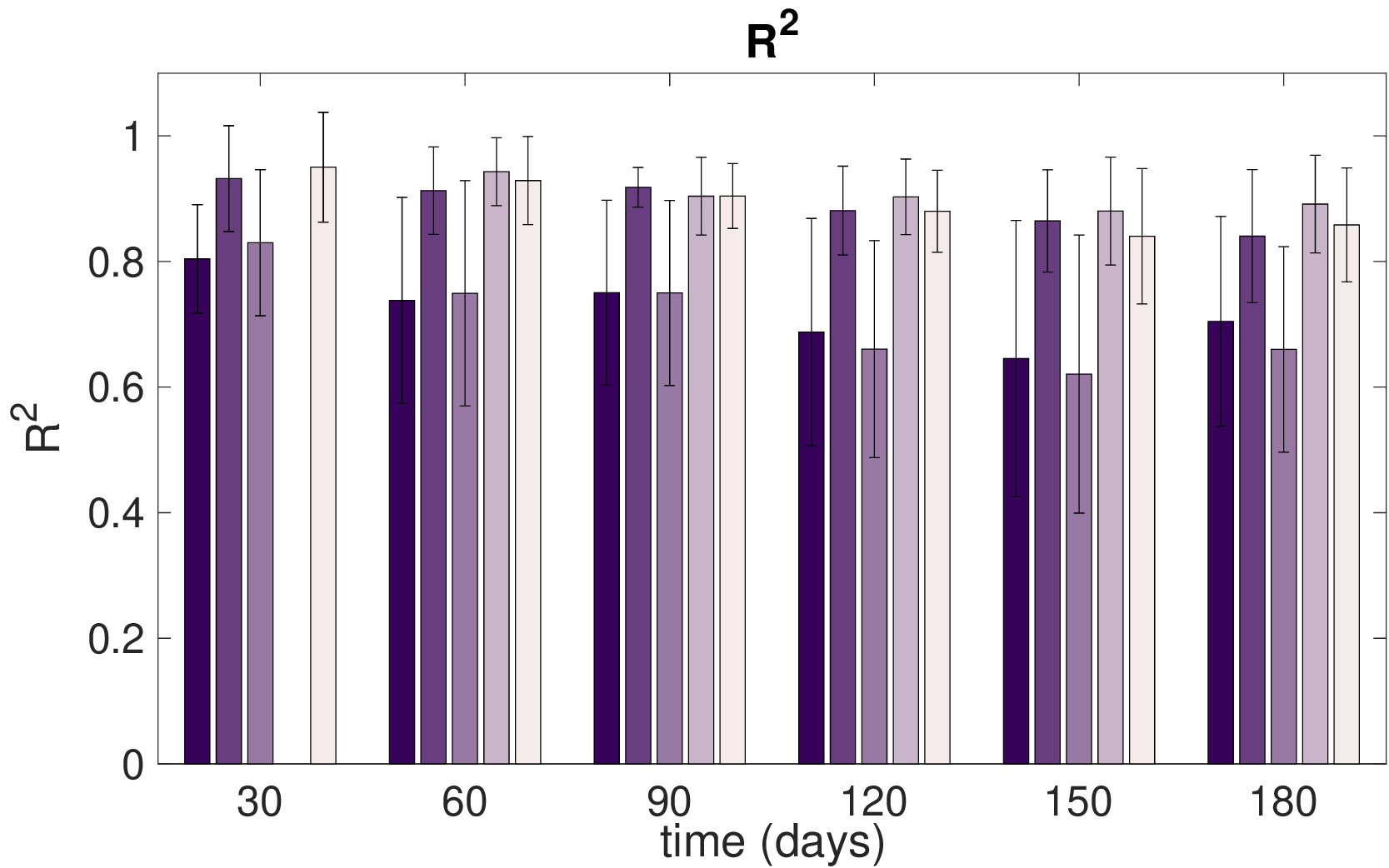}\label{fig:R2}}
                    \subfigure[]{\includegraphics[clip,width=0.49\columnwidth, height=0.30625\columnwidth]{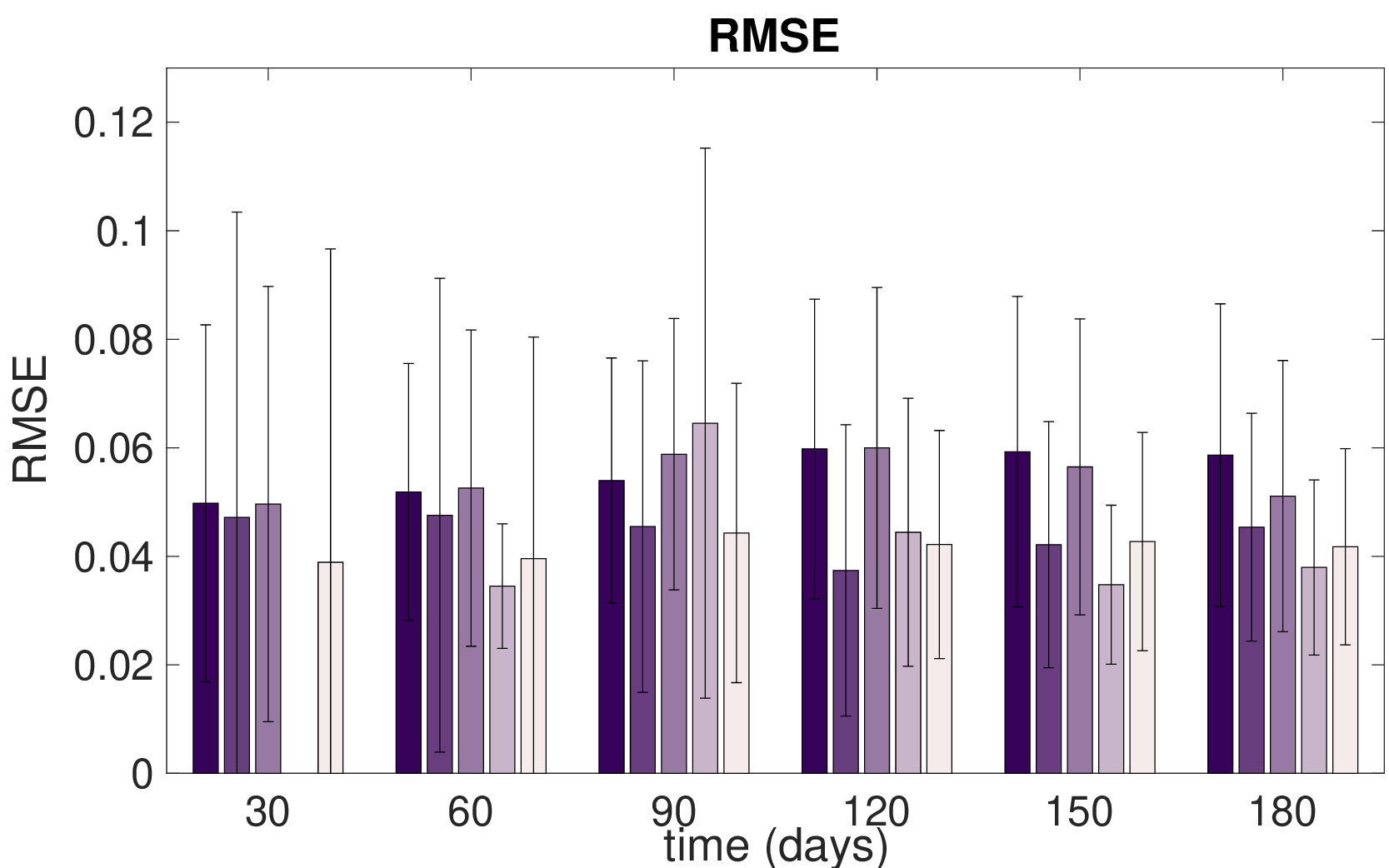}\label{fig:RMSE}}
                           \label{fig:}
                    \subfigure[]{\includegraphics[clip,width=0.49\columnwidth, height=0.30625\columnwidth]{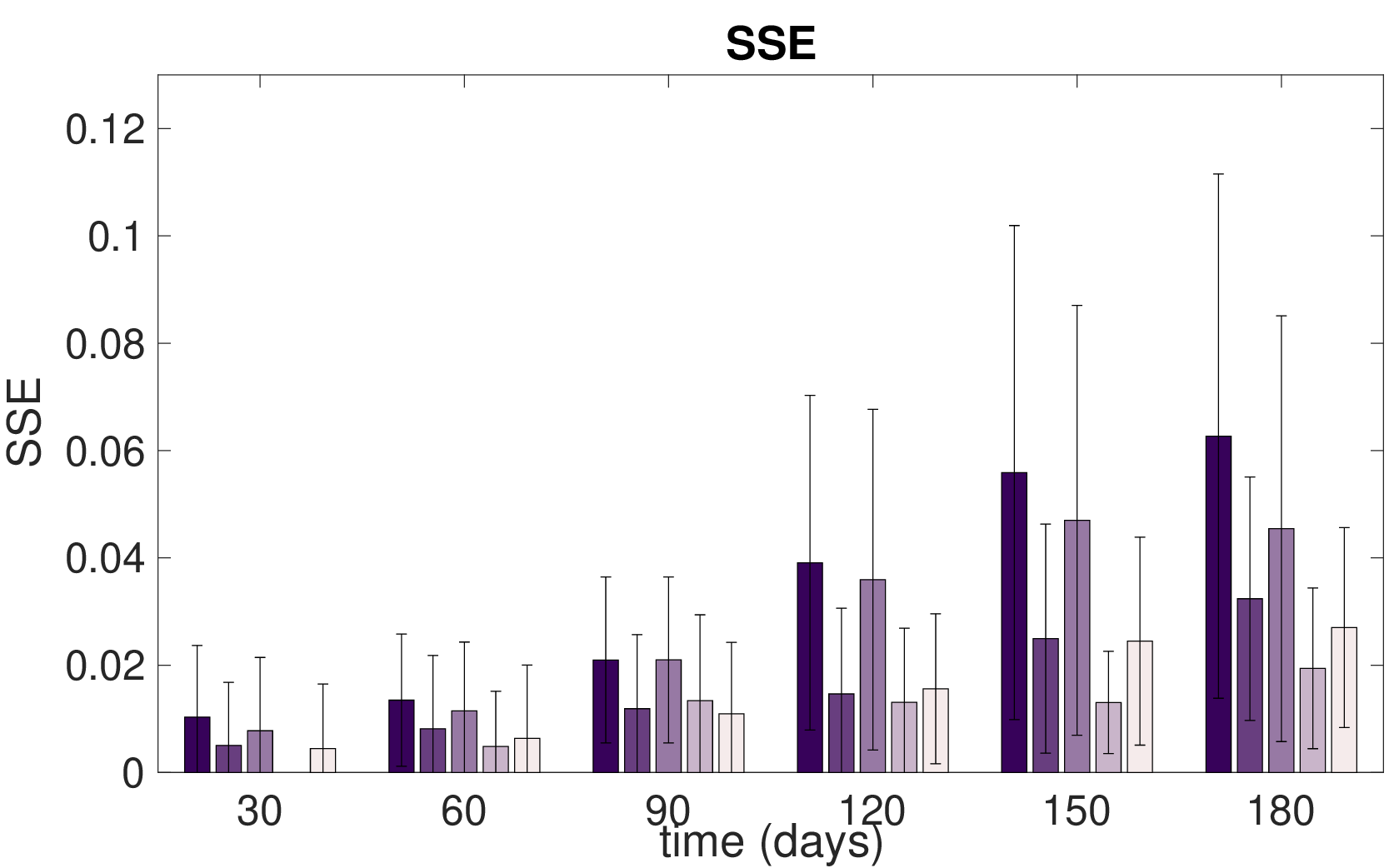}\label{fig:SSE}}
                     \subfigure[]{\includegraphics[clip,width=0.49\columnwidth, height=0.30625\columnwidth]{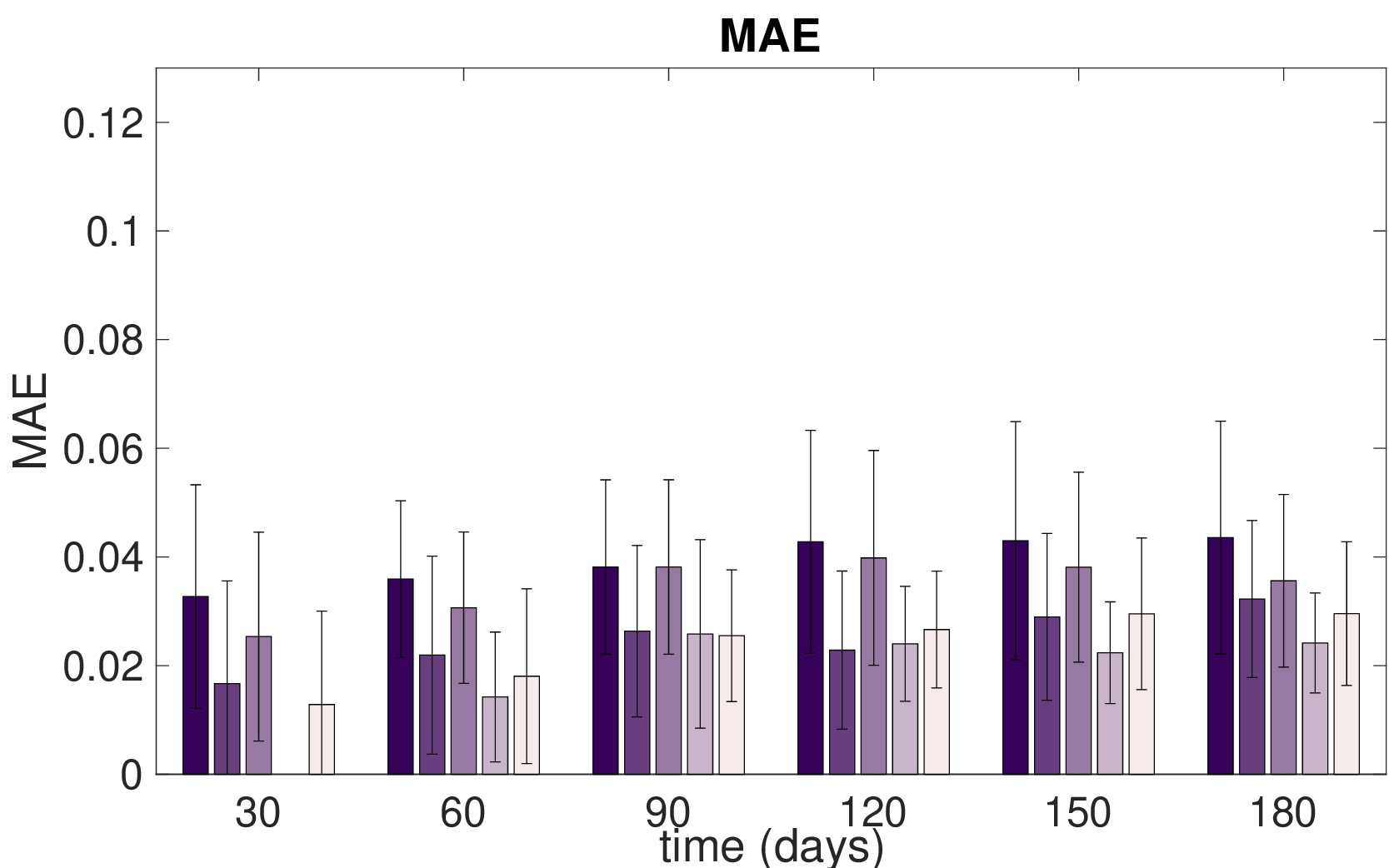}\label{fig:MAE}}
            
                    \caption{Accuracy metrics of the {LS}-supported Bayesian regression fitting over time. Five different models are evaluated: $exp1$~\textcolor{morado1}{$\blacksquare$}, $exp2$~\textcolor{morado2}{$\blacksquare$}, $gauss1$~\textcolor{morado3}{$\blacksquare$}, $gauss2$~\textcolor{morado4}{$\blacksquare$}, $poly3$~\textcolor{morado5}{$\blacksquare$}.~(\textbf{a}) Coefficient of determination, R$^2$.~(\textbf{b}) Root mean squared error, RMSE.~(\textbf{c}) Sum of squared errors, SSE.~(\textbf{d}) Mean absolute error, MAE.}  
                    \label{fig:BayFig}
            
                    \end{figure}

Figure~\ref{fig:BayFig} presents the accuracy metrics of the five Bayesian regression models introduced in Table~\ref{tab:table2} to estimate the performance of PCE over 180 days, for each of the OSCs available in the database (Table~\ref{tab:table1}). It is worth noting that these results correspond to the mean values, along with the standard deviation for each OSC. {In general terms, it can be observed that these models tend to perform slightly better over short time periods {compared to the longer experiments.}}

%

The maximum errors for all temporal fittings are bounded to values~$\sim$0.06. According to the results of Figure~\ref{fig:BayFig}, it is observed that on average, the $gauss2$ model, consisting of two Gaussian terms, is the one that best estimates the performance of the OSCs in terms of PCE over time. However, it should be noted that for a 30{-}day fit, this model requires more points to achieve a valid R$^2$.  
{It is worth mentioning that} these methods {cannot} characterize the behavior of all devices contained in our database in a global manner, {and the} results {obtained} reflect solely the mean performance of fitting for {individual} OSCs.
\clearpage        
{\subsubsection{PCE prediction with LS}}       

\begin{figure}[h!]
                \centering
                \includegraphics[scale=0.45,clip]{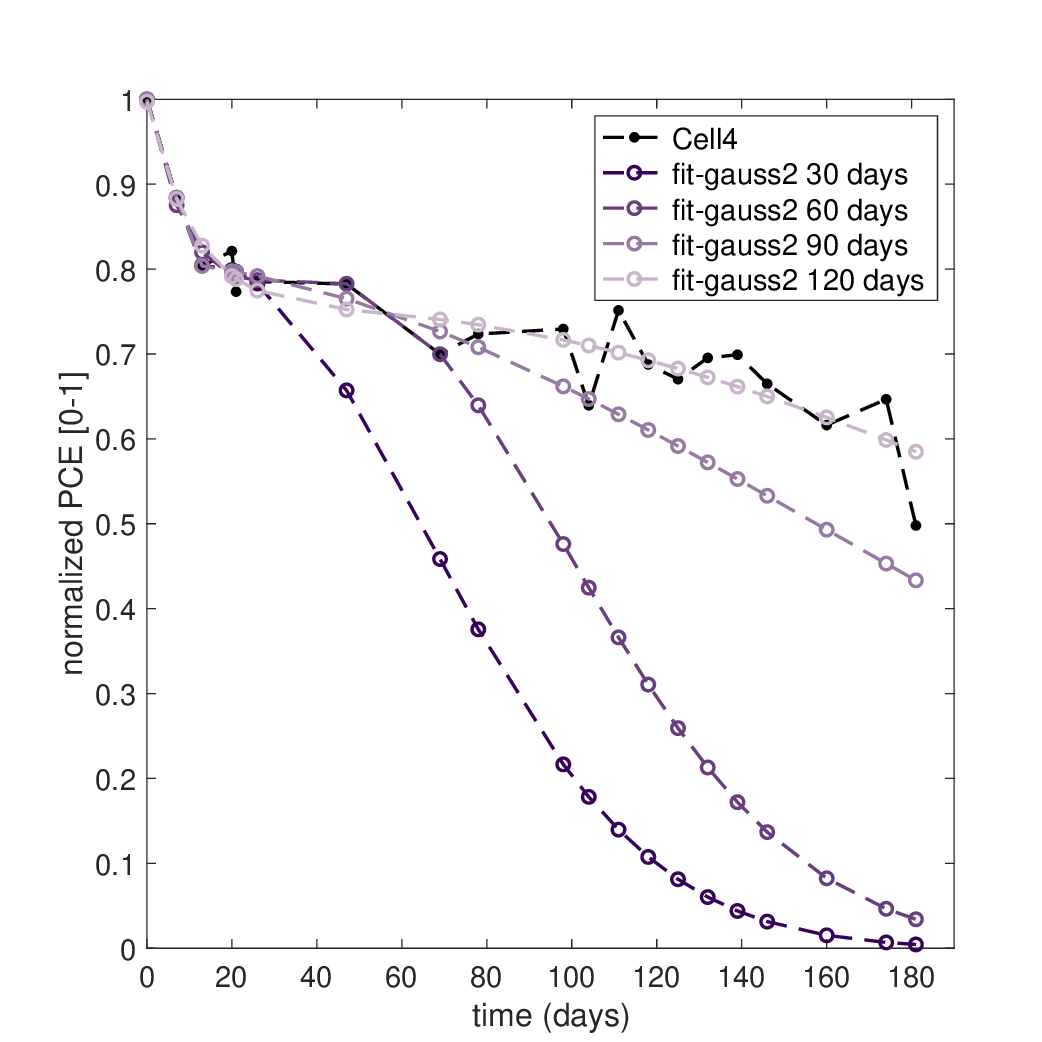}
               \caption{PCE {predition} over time with the best {LS}-supported Bayesian regression fitting model ($gauss2$). Four temporal datasets are used to compute the fittings: 30 days~\textcolor{violeta1}{$-\circ-$}; 60 days~\textcolor{violeta2}{$-\circ-$}; 90 days~\textcolor{violeta3}{$-\circ-$} and 120 days~\textcolor{violeta4}{$-\circ-$}.}
               \label{fig:BayPredict}
            \end{figure}

Once preliminary fitting{s} have been made with traditional {LS} regression methods, it has been found that the best fitting corresponds to a double-term Gaussian function. Now we analyze the robustness of this method to predict the behavior of PCE against the temporal variable. For this purpose, {four regression fittings have been obtained with} PCE data acquired up to: 30, 60, 90, and 120 days, respectively. Then the behavior of PCE at 180 days has been predicted {for each one} (Figure~\ref{fig:BayPredict} {and Table~\ref{tab:table3})}. It can be {confirmed that only the fitting with data up to 120 days is able to predict reliably future PCE values}. This {result} demonstrates that even the best parametric {LS} fitting regression model requires more than half of the temporal data to reliably model the PCE evolution at future time values.
            \begin{table}[h!]
                    \caption{Accuracy metrics of the best {LS}-supported Bayesian regression fitting model ($gauss2$) to {predict} PCE over time, presented in Figure~\ref{fig:BayPredict}.}
                    \label{tab:table3}
                    \centering
                    \footnotesize
                    \begin{tabular}{r|rccc}
                    \toprule
                    \textbf{Fitting data}&\textbf{{Prediction}}&\textbf{RMSE}&\textbf{SSE}&\textbf{MAE}\\
                    \midrule
                    30 days&60 days&0.0516&0.0213&0.0277\\ 
                   		&90 days&0.1141&0.1302&0.0680\\ 
                   		&120 days&0.1872&0.4905&0.1339\\ 
            		&150 days&0.2224&0.8405&0.1705\\ 
            		&180 days&0.2442&1.3117&0.1979\\ 
                    \hline
                60 days&90 days&0.0454&0.0206&0.0220\\ 
                   		&120 days&0.1083&0.1641&0.0685\\ 
            		&150 days&0.1511&0.3879&0.1046\\ 
            		&180 days&0.1911&0.8038&0.1445\\ 
                    \hline
                    90 days
                   		&120 days&0.0150&0.0032&0.0103\\ 
            		&150 days&0.0160&0.0043&0.0119\\ 
            		&180 days&0.0190&0.0079&0.0118\\ 
                    \hline
                    120 days
                    		&150 days&0.0158&0.0042&0.0127\\ 
            		&180 days&0.0138&0.0038&0.0105\\          
                    \bottomrule
                    \end{tabular}
                    \end{table}

\pagebreak
\subsection{{ML} regression results}

  \begin{table}[]
                    \caption{Accuracy metrics of the ML regression models computed with training data up to 180 days.}
                    \label{tab:table9}
                    \centering
                    \footnotesize
             \begin{tabular}{r|rcccc}
                    \toprule
                    \textbf{Training data}&\textbf{ML model}&\textbf{R$^2$}&\textbf{RMSE}&\textbf{SSE}&\textbf{MAE}\\
                    \midrule
                    180 days
                    		&GB-90-10&0.96&0.0640&0.0595&0.0520\\
            		&GB-80-20&0.96&0.0720&	0.0600&	0.0560\\
                   		&GB-70-30&0.94&0.0840&	0.0672&	0.0650\\
            		&GB-60-40&0.95&0.0700&	0.0431&	0.0500\\
            		\cline{2-6} 
                   		&NN-90-10&0.66&0.1400&	0.3732&	0.1100\\
            		&NN-80-20&0.76&0.1100&	0.1138&	0.0670\\
            		&NN-70-30&0.60&0.1600&	0.3555&	0.1200\\
            		&NN-60-40&0.43&0.2500&	0.7265&	0.1800\\
            		\cline{2-6} 
            		&MVL-90-10&0.81&0.1000&	0.2929&	0.0810\\
            		&MVL-80-20&0.81&0.1000&	0.2783&	0.0800\\
            		&MVL-70-30&0.74&0.1400&	0.2696&	0.1000\\
            		&MVL-60-40&0.74&0.1300&	0.2802&	0.0920\\
            		\cline{2-6} 
            		&\textbf{RF-90-10}&\textbf{0.96}&\textbf{0.0480}&	\textbf{0.0246}&	\textbf{0.0390}\\
            		&RF-80-20&0.97&0.0520&	0.0309&	0.0370\\
                   		&RF-70-30&0.94&0.0670&	0.0398&	0.0460\\
                   		&RF-60-40&0.93&0.0850&	0.0644&	0.0580\\
                    \bottomrule
                    \end{tabular}
                    \end{table}

In this section, we {generate} ML models to {study} the temporal behavior of PCE. Notice that, unlike the previous {LS}-based regression fittings presented in Section~\ref{sec:bay}, these ML models allow us to characterize the performance of {all the OSC} devices {under a single model, while also using the multiple variables in the dataset. These include} manufacturing {parameters and} environmental conditions at each measurement. {To} comparatively assess the  ML models, {Table~\ref{tab:table9}} presents detailed accuracy results for the  algorithms GB, NN, MVL, and RF, {with training data up to 180 days. {Please note that all these models have PFI enabled, so that the most important features are considered. For further details about these features see Section~\ref{sec:features} and \ref{sec:app3}}.

The nomenclature for the models is: ALGORITHM-TRAINING-VALID (e.g.~RF-90-10). Results with different training-validation ratios are presented: from 90-10\% to 60-40\%. {In this study, the optimal ML models used RF algorithms with a 90-10 training-validation partitioning.} It can be observed that, in general, all {MAE and RMSE values} are bounded within ranges $\sim$[0.02-0.03], for training data of 120 days onwards (Figure~\ref{fig:ML}).

The most robust model is highlighted in bold, considering the overall performance across all presented metrics, with priority given to the R$^2$ value. {Please note that the supplementary material (\ref{sec:app}) contains these same metrics for further inspection of models obtained with other temporal ranges.} It is worth noting that errors of the ML methods are clearly bounded, {regardless they operate on a} multivariate database, in contrast to univariate Bayesian fitting {presented in Section~\ref{sec:bay}}.

                    \begin{figure}[h!]
                    \centering
                            \subfigure[]{\includegraphics[clip,width=0.49\columnwidth, height=0.416\columnwidth]{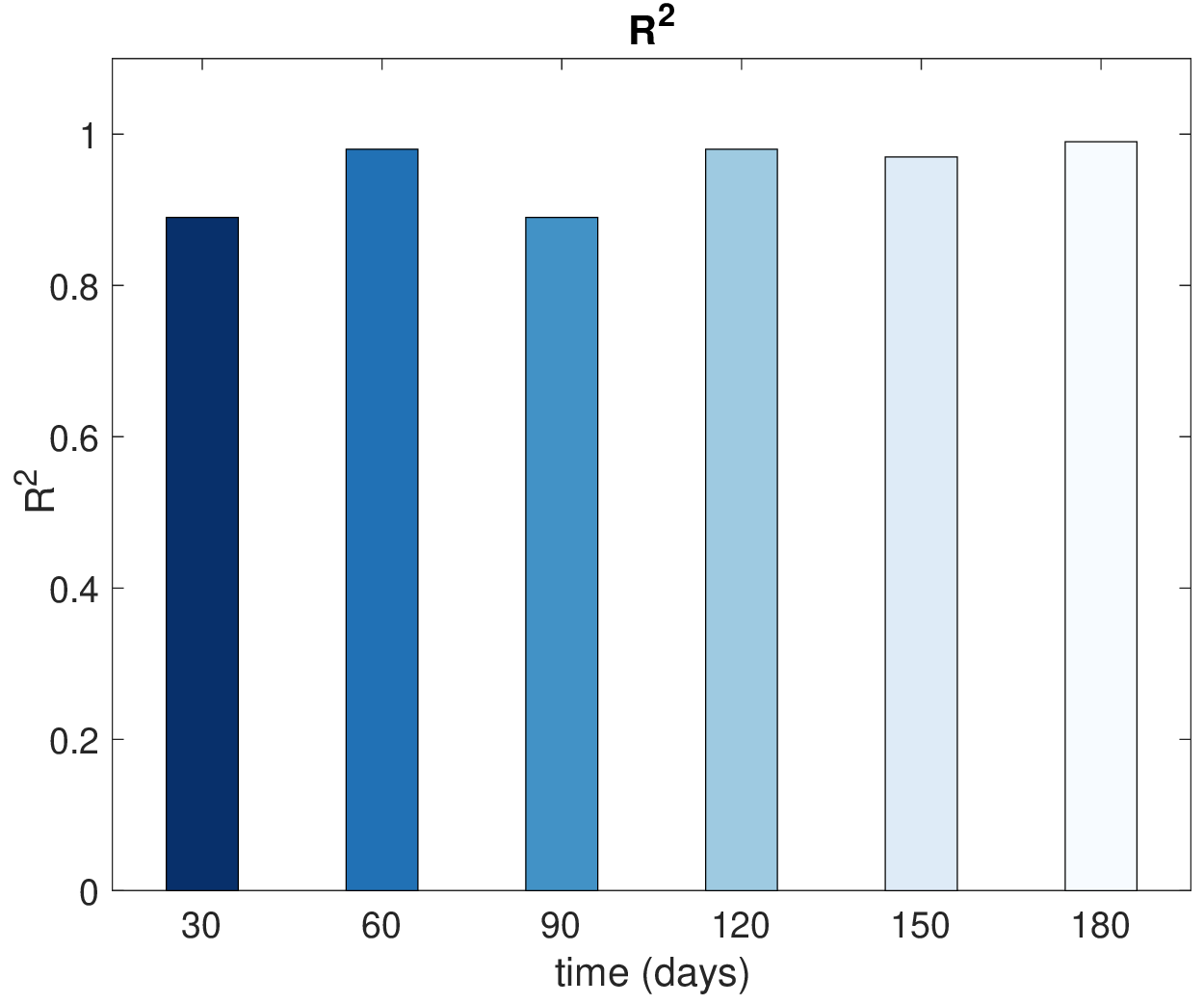}\label{fig:ML1}}
                    \subfigure[]{\includegraphics[clip,width=0.49\columnwidth, height=0.416\columnwidth]{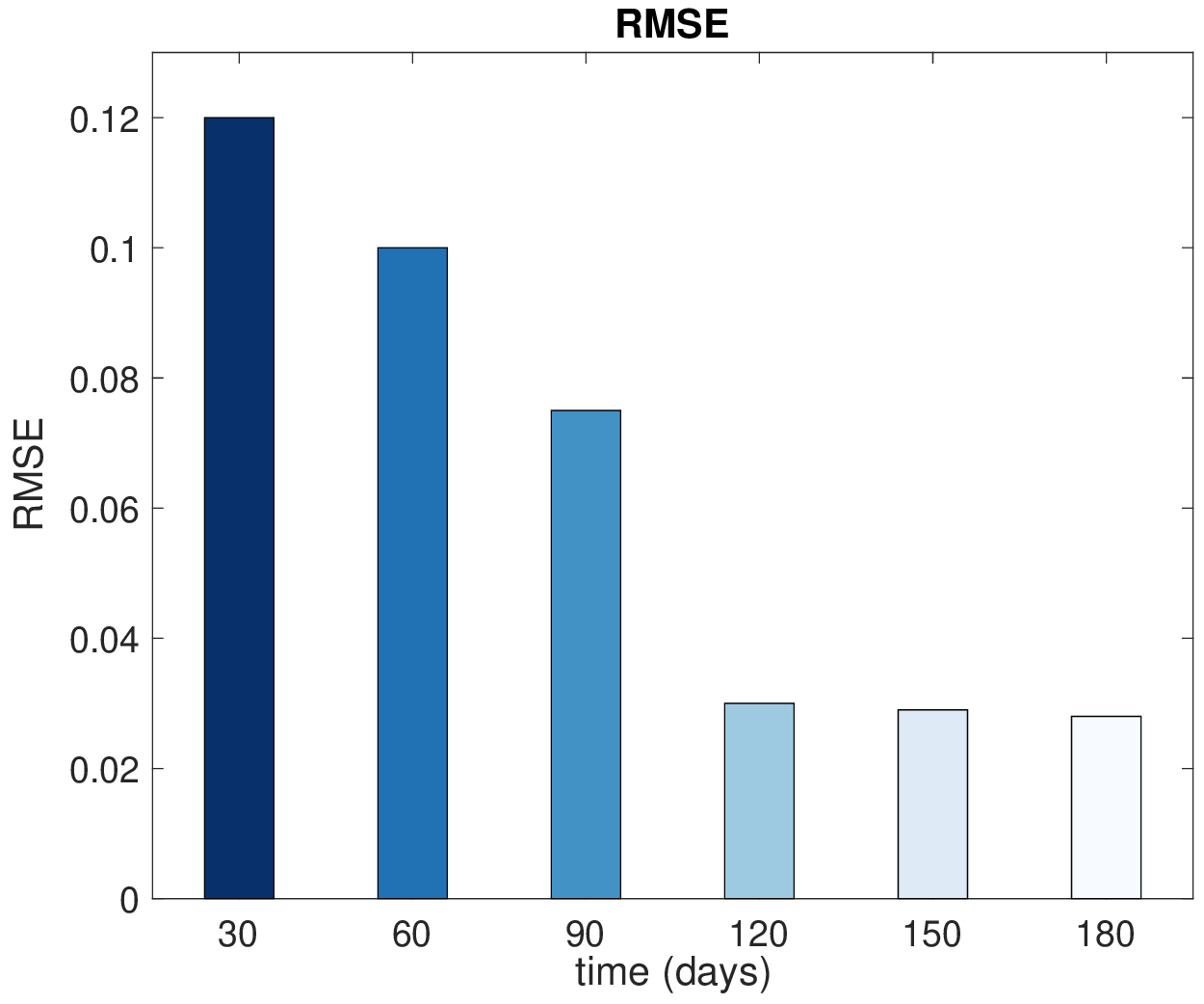}\label{fig:ML2}}
                      
                    \subfigure[]{\includegraphics[clip,width=0.49\columnwidth, height=0.416\columnwidth]{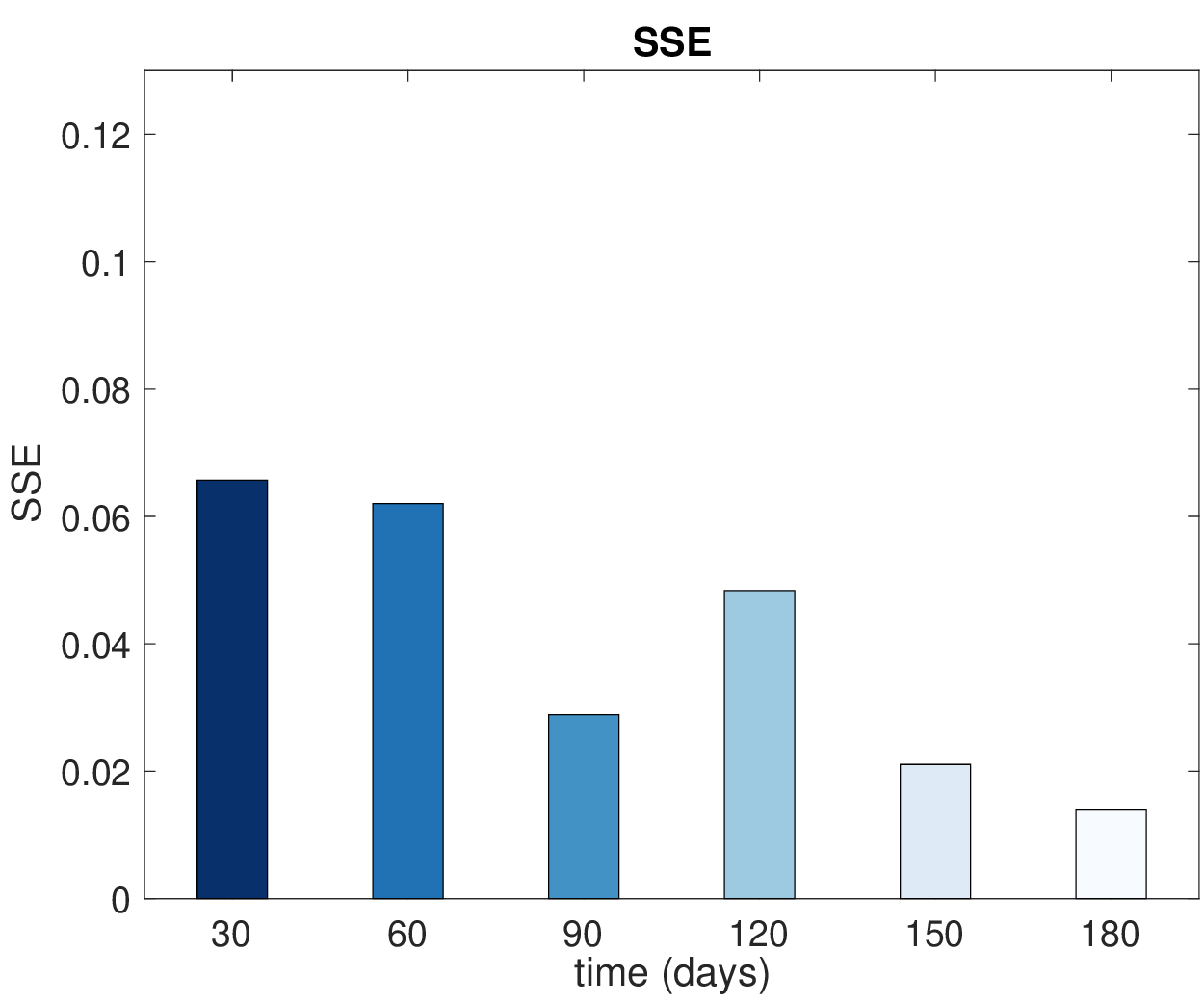}\label{fig:ML3}} 
                     \subfigure[]{\includegraphics[clip,width=0.49\columnwidth, height=0.416\columnwidth]{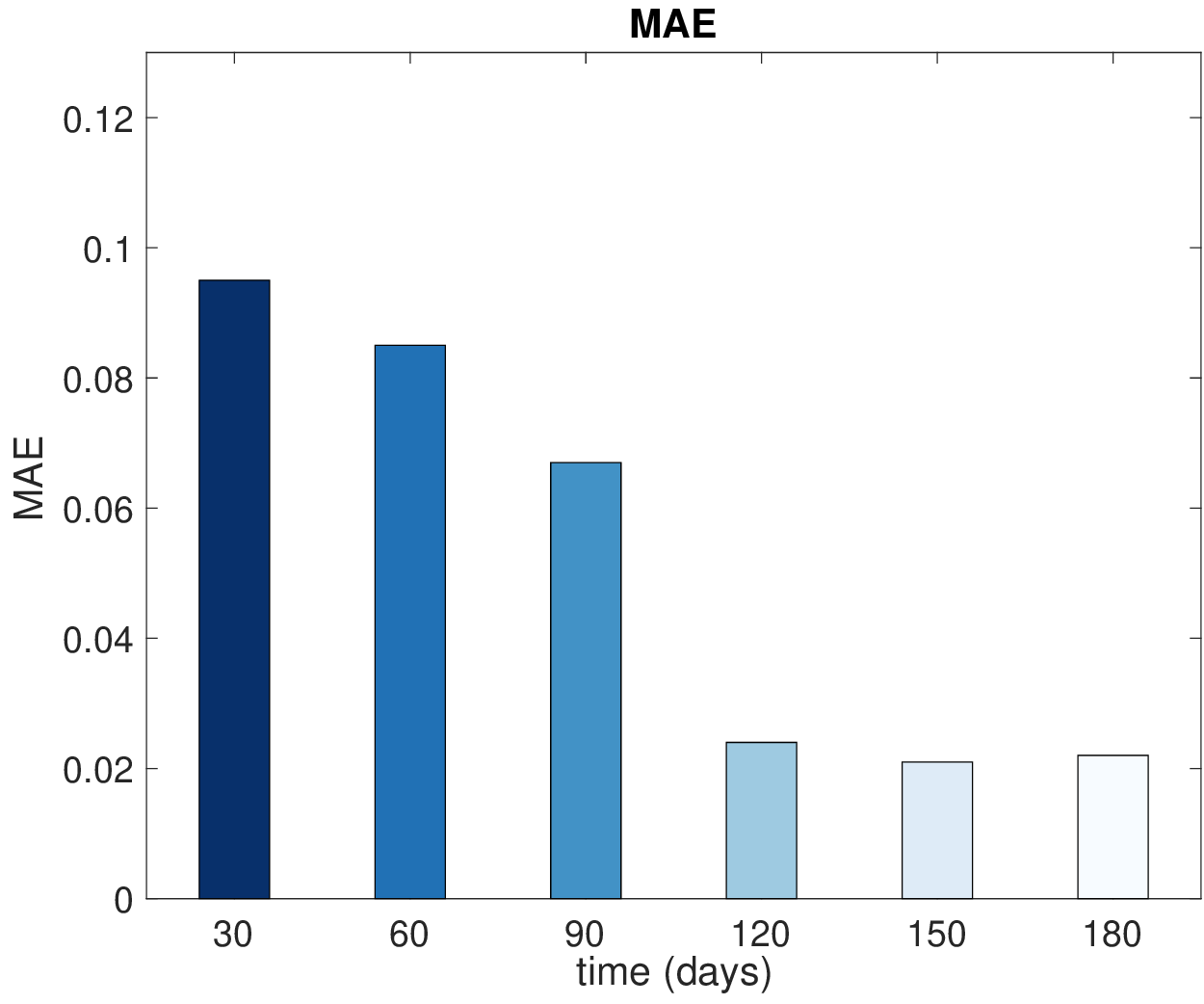}\label{fig:ML4}}
            
                    \caption{Accuracy of the best {ML} regression models over time: 30 days RF-60-40~\textcolor{azul1}{$\blacksquare$}; 60 days RF-90-10~\textcolor{azul2}{$\blacksquare$}; 90 days RF-90-10~\textcolor{azul3}{$\blacksquare$}; 120 days RF-90-10~\textcolor{azul4}{$\blacksquare$}; 150 days RF-90-10~\textcolor{azul5}{$\blacksquare$} and 180 days RB-90-10~\textcolor{azul6}{$\blacksquare$}.~(\textbf{a}) Coefficient of determination, R$^2$.~(\textbf{b}) Root mean squared error, RMSE.~(\textbf{c}) Sum of squared errors, SSE.~(\textbf{d}) Mean absolute error, MAE.} 
                    \label{fig:ML}
            
                    \end{figure}

          \clearpage
\subsubsection{PCE prediction {with ML}\label{sec:preML}}

After comparatively evaluating the performance of ML models, this section validates their ability to predict the temporal behavior of the PCE for an OSC device never seen by the models. Firstly, the models are trained and validated without data of $Cell4$. Standing out above the others are the GB-90-10 and RF-90-10 models. {In line with previous results (Table~\ref{tab:table9}), and thus considering its robustness to model the entire dataset, RF-90-10 has been selected to get} predictive inferences up to 180 days for the unseen OSC $Cell4$. Figure~\ref{fig:results} provides such results. {In particular, Figure~\ref{fig:noC4} presents the results of the model when the dataset is splitted into traning and validation, at 90\%-10\% (blue and yellow points), whereas Figure~\ref{fig:preC4} shows the prediction for $Cell4$ when its data are used as external test.}

Next, Figure~\ref{fig:cross} presents validation tests. Figure~\ref{fig:test} shows the RMSE error of the model compared to those of the tests: y-mean, y-shuffle and onehot. Figure~\ref{fig:CV} produces a 5-fold cross-validation test~\cite{CV5F}, with satisfactory results. In consequence, neither data leakage nor overfitting is evidenced.



    \begin{figure}[h!]
                    \centering
 \subfigure[]{\includegraphics[clip,scale=0.35]{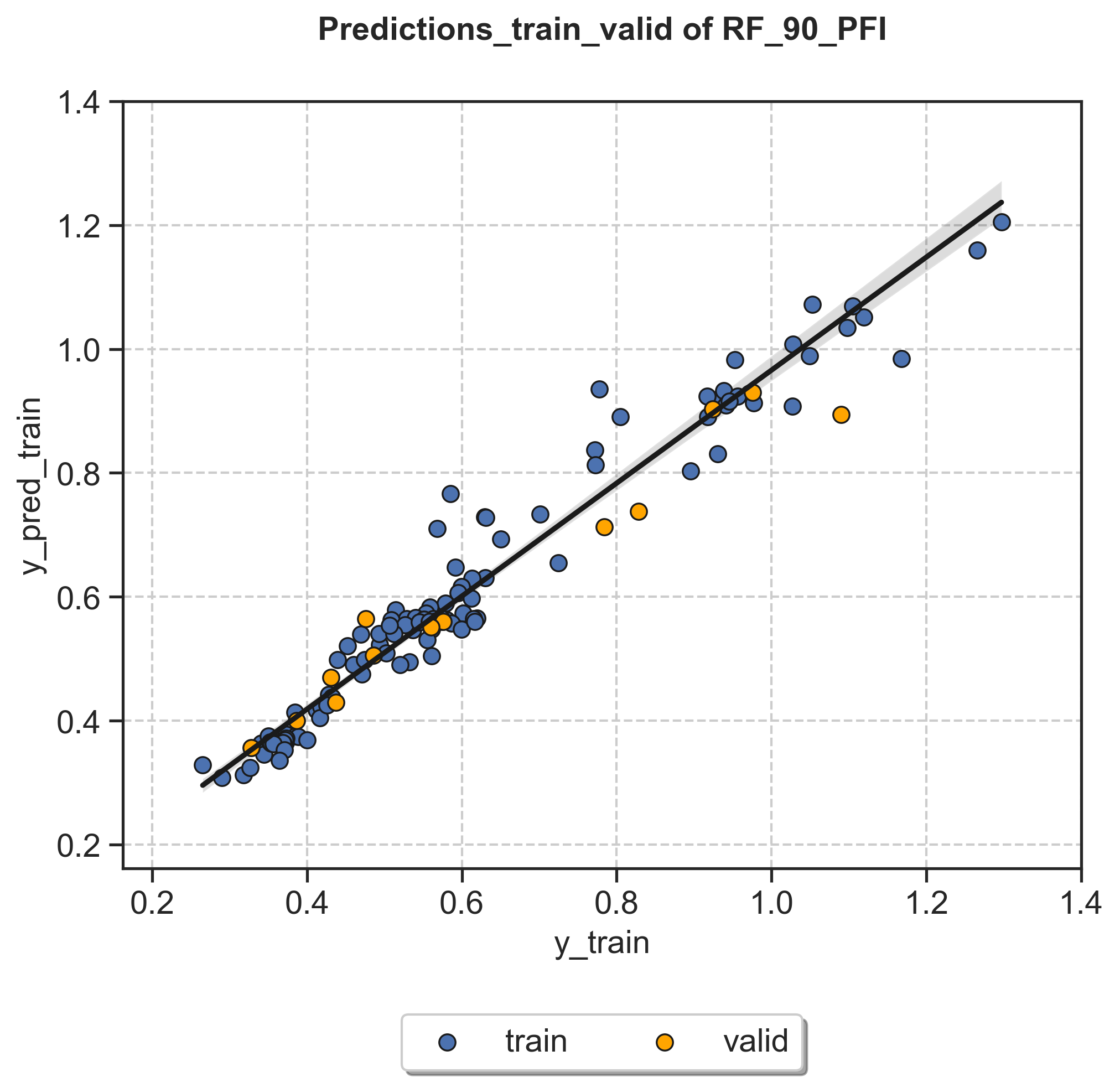}\label{fig:noC4}}                   
                    \subfigure[]{\includegraphics[clip,scale=0.35]{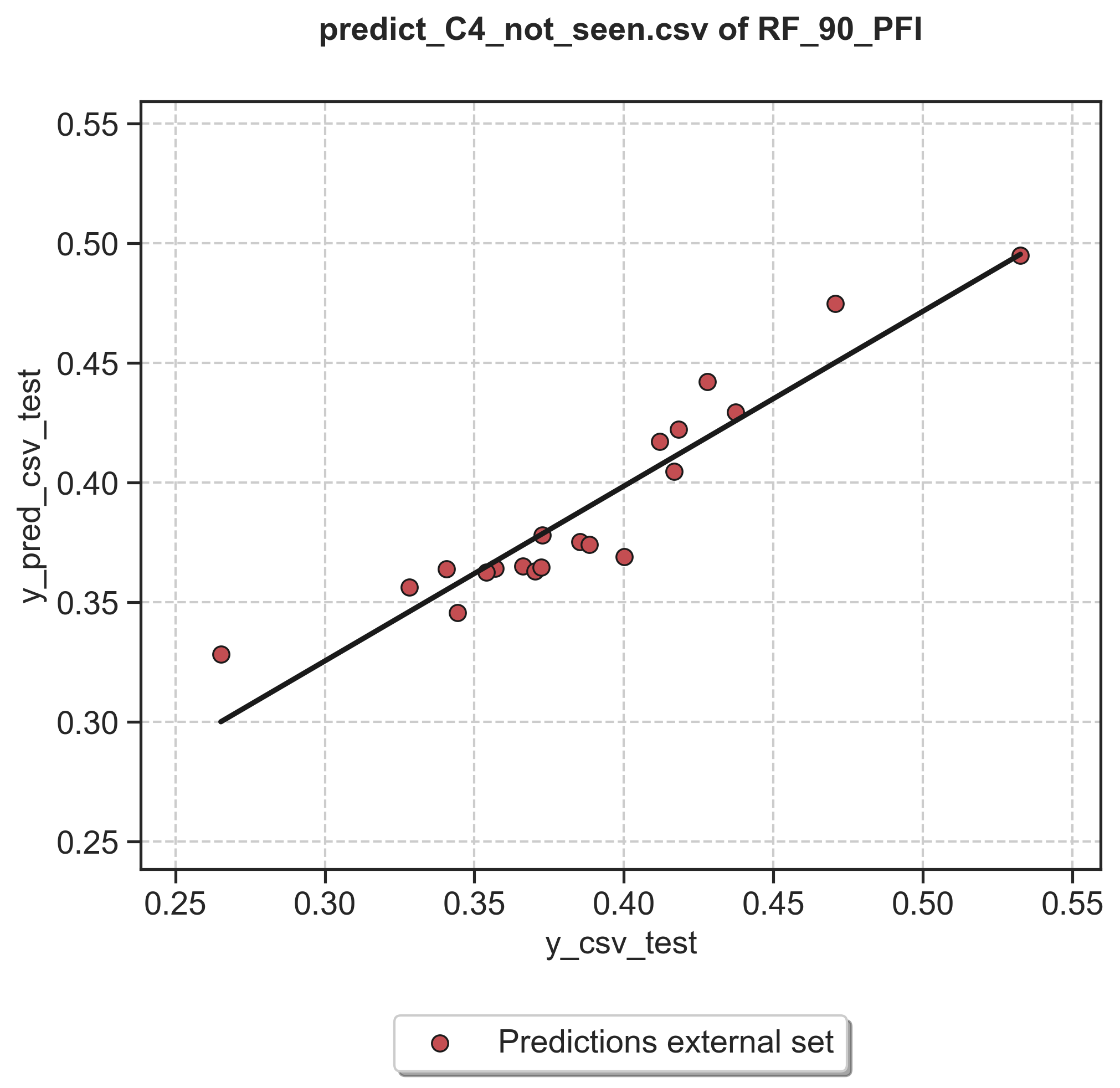}\label{fig:preC4}}
                  
                    \caption{{Results of the best ML model, RF-90-10:~(\textbf{a}) training-validation without data of $Cell4$ (R$^2$=0.96, MAE=0.05, RMSE=0.071);~(\textbf{b}) prediction of $Cell4$ as external test (R$^2$=0.88, MAE=0.015, RMSE=0.021).}}  
 \label{fig:results}
\end{figure}
 
 \pagebreak
 \begin{figure}[h!]
                    \centering
 \subfigure[]{\includegraphics[clip,scale=0.35]{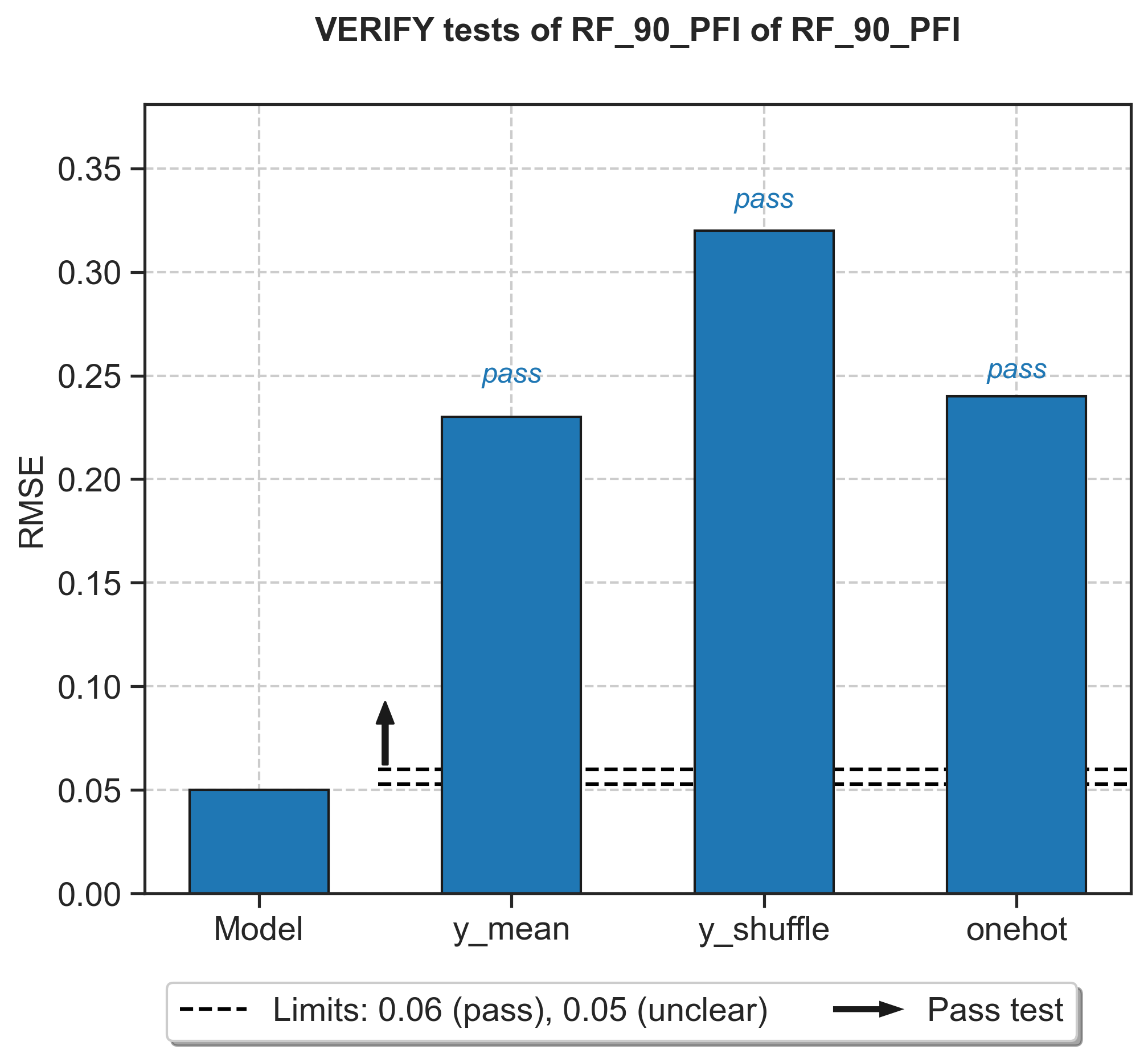}\label{fig:test}}                   
                    \subfigure[]{\includegraphics[clip,scale=0.35]{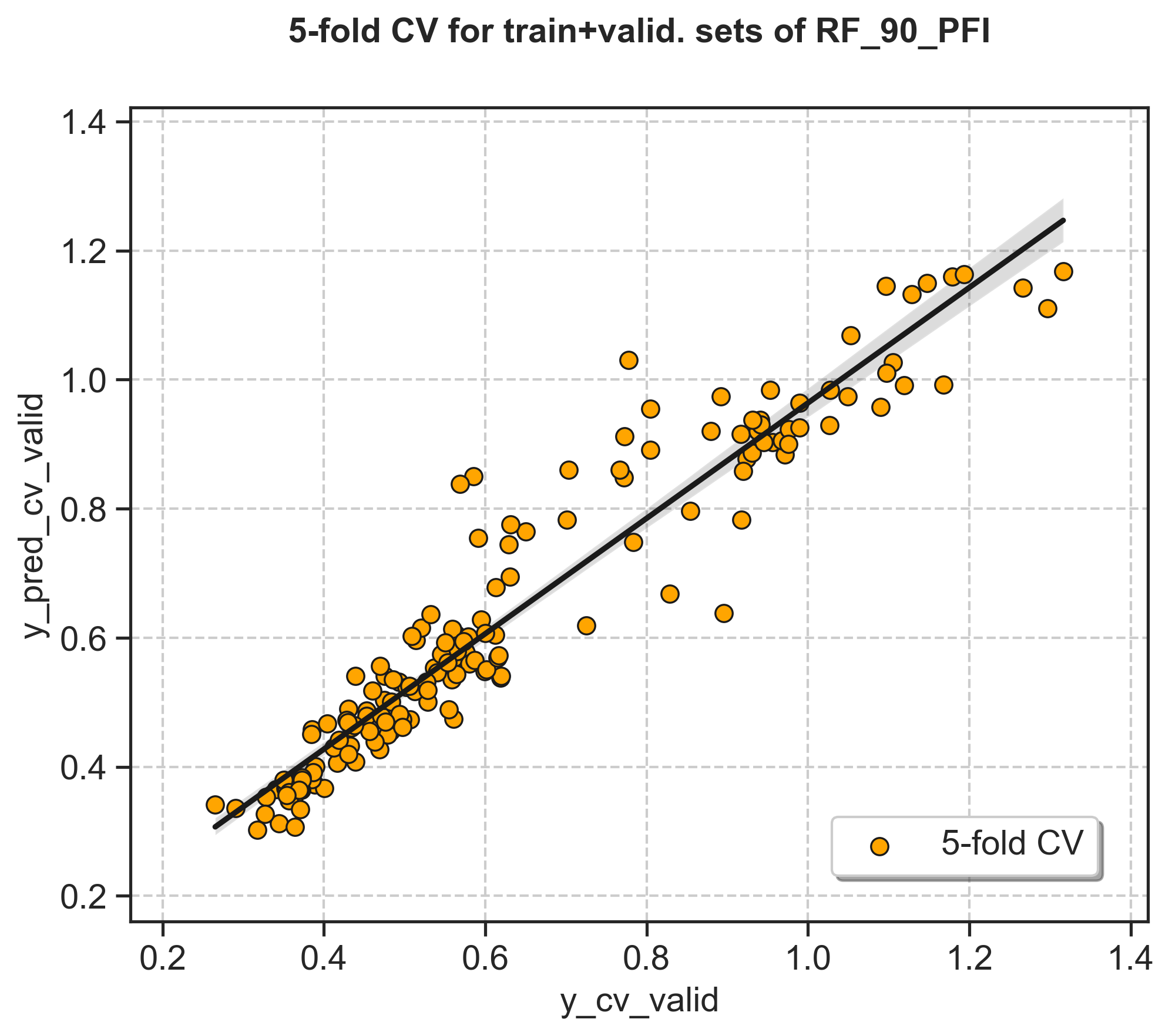}\label{fig:CV}}
                  
                    \caption{{Validation of the best ML model, RF-90-10:~(\textbf{a}) RMSE values for validation tests y-mean, y-shuffle and onehot;~(\textbf{b}) 5-fold cross-validation test (R$^2$=0.89, MAE=0.06, RMSE=0.09).}}  
 \label{fig:cross}
\end{figure}
 
Finally, Figure~\ref{fig:MLPre} compares the temporal evolution of the PCE for $Cell4$ with the predicted data obtained by the model.
            \begin{figure}[h!]
                \centering
                \includegraphics[scale=0.45,clip]{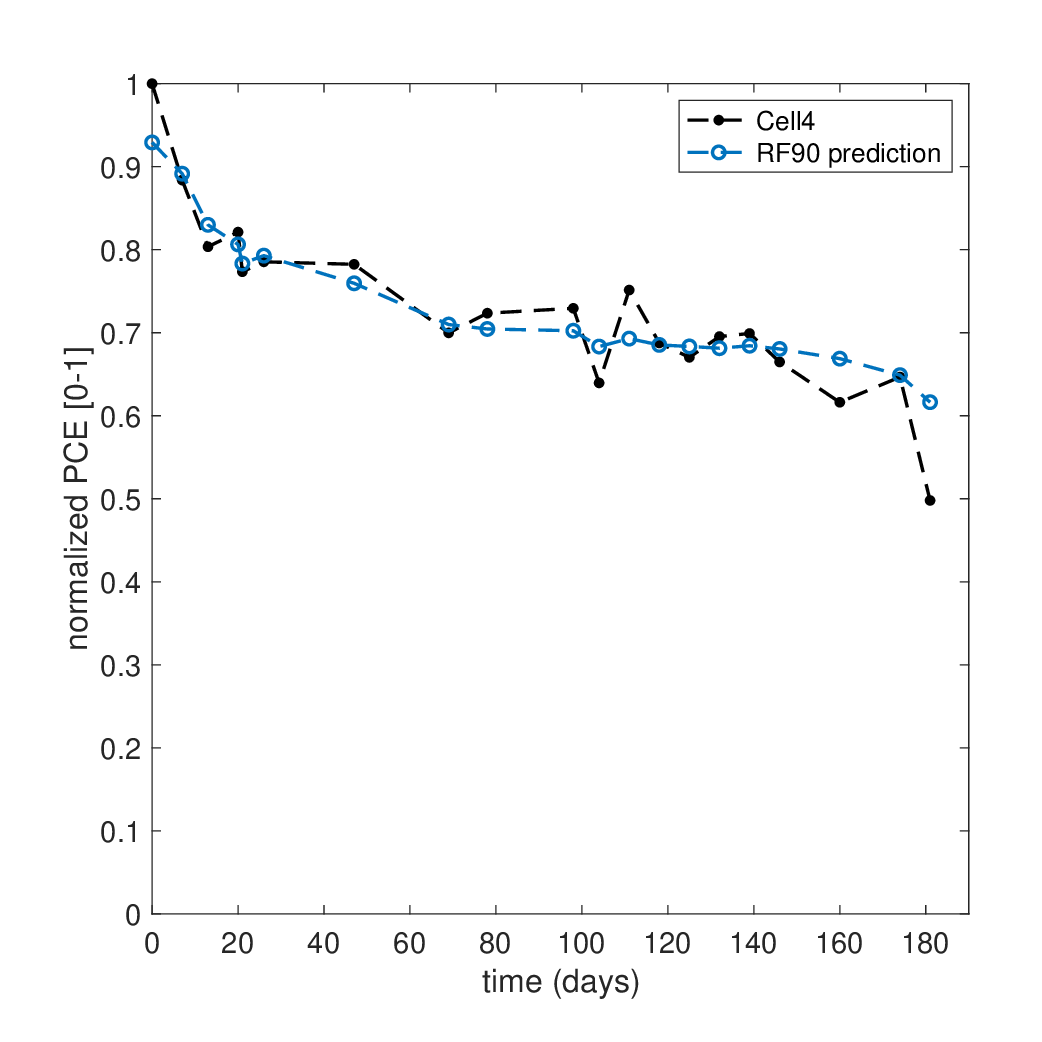}
               \caption{PCE prediction over time with the best {ML} regression model (RF-90-10, blue) for an OSC not seen during the training-validation process ($Cell4$, black).}
               \label{fig:MLPre}
            \end{figure}
\clearpage
 \subsubsection{{Feature analysis}}\label{sec:features}

{Considering} the results obtained in the previous section, it is worthwhile to study the top-performing ML model that characterizes the behavior of OSCs (i.e. RF-90-10). {ML models trained with a limited number of data points are unable to characterize and predict with sufficient accuracy. For this reason, we focused on analyzing the complete dataset, with measures up to 180 days.} In this regard, PFI and SHAP analyses are presented below, by means of Figure~\ref{fig:PFI} and Figure~\ref{fig:SHAP}, respectively. {PFI evaluates the significance of individual features in a ML model by measuring the increase in the model's prediction error after permuting the values of a specific feature. The resulting increase in error indicates the dependency on that feature~\cite{PFI1,PFI2}. Besides, SHAP permits interpreting individual predictions of ML models by computing the contribution of each feature to the model's as an additive feature attribution method~\cite{SHAP2017}}.

Then Figure~\ref{fig:PFI} presents the influence of the most relevant feature for the model, after applying PFI filtering, that is, removing variables with low effect on the R$^2$. It is confirmed that the amount of solvent in the HTL layer, i.e. the amount of PEDOT:PSS, has the most significant influence. Likewise, the P3HT:PCBM ratio also demonstrates certain relevance. Additionally, dependencies with the value of PCBM are also observed, since it constraints the P3HT:PCBM ratio, given its non-linear influence in the denominator. 

As for the environmental conditions, it is known that low humidity proves to be beneficial for these OSCs, which in our geographical location is normally correlated with high temperatures and atmospheric pressures. In a similar manner it acts the dew point, which directly correlates with humidity. Nonetheless, the device encapsulation demonstrates that the effect of these variables is minimized, as PFI filtering proves that they are not relevant enough to influence substantially the model.
 \begin{figure}[h!]
                    \centering
                    {\includegraphics[clip,scale=0.4]{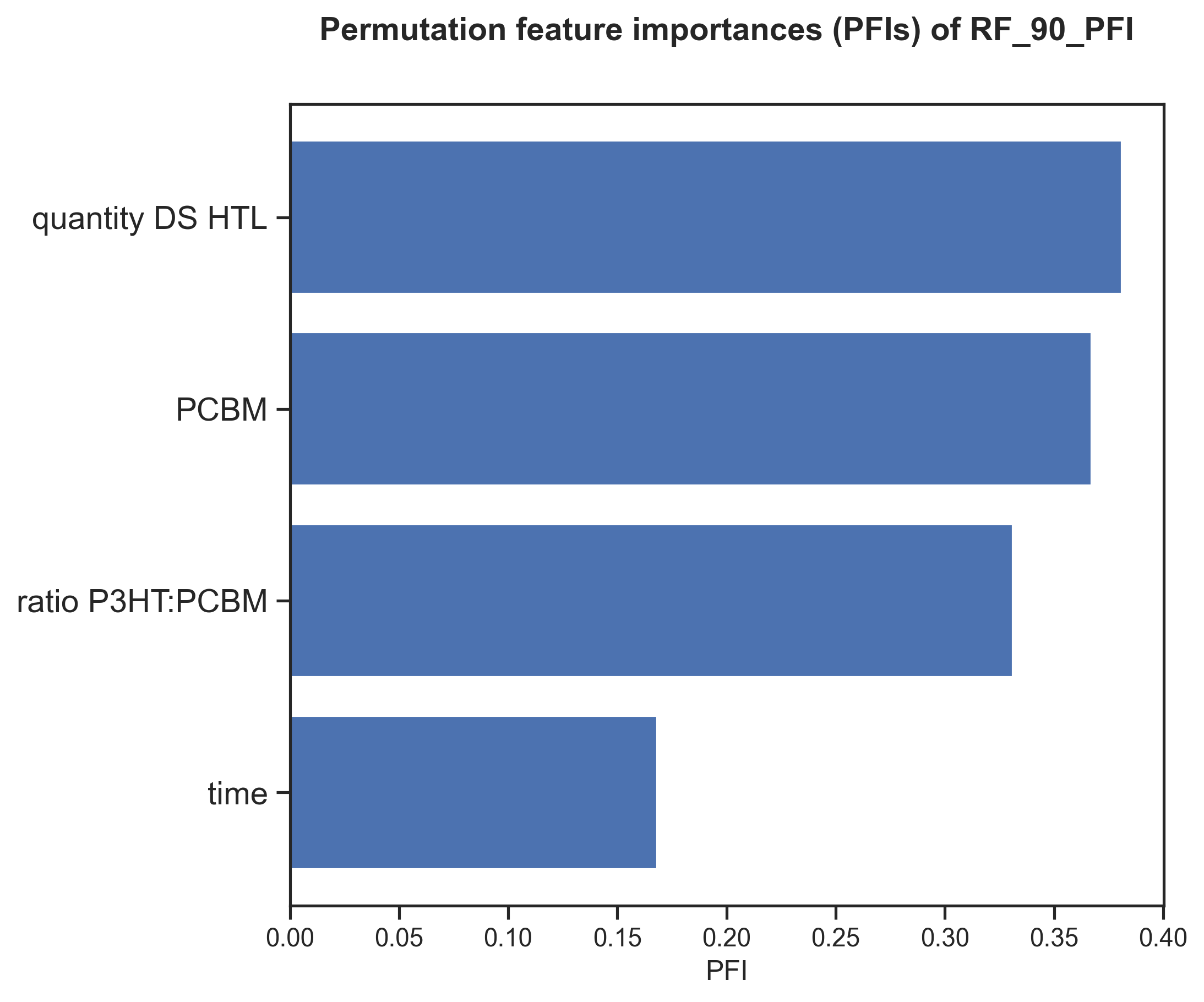}\label{fig:pfi4}}  
                                 {    \caption{PFI results for the best {ML} model generated with data up to 180 days.}  \label{fig:PFI}}
                    
\end{figure}

{Regarding the SHAP analysis, Figure~\ref{fig:SHAP} validates the previous insights: high values of the amount of solvent PEDOT:PSS in the HTL layer, high ratios of P3HT:PCBM are relevant to get stable PCE values. Besides this, the effect of the PCBM value on the P3HT:PCBM ratio is again confirmed.


{Overall, the importance of PEDOT:PSS has been demonstrated as it plays an essential role in the multilayer structure of the OSCs. It confirms its relevance as the second layer to cover sufficiently the substrate.~Moreover, the ratio P3HT:PCBM also demonstrates its role in dealing as charge carrier in the HTL layer.}

%
\begin{figure}[h!]
                    \centering
                    {\includegraphics[clip,scale=0.45]{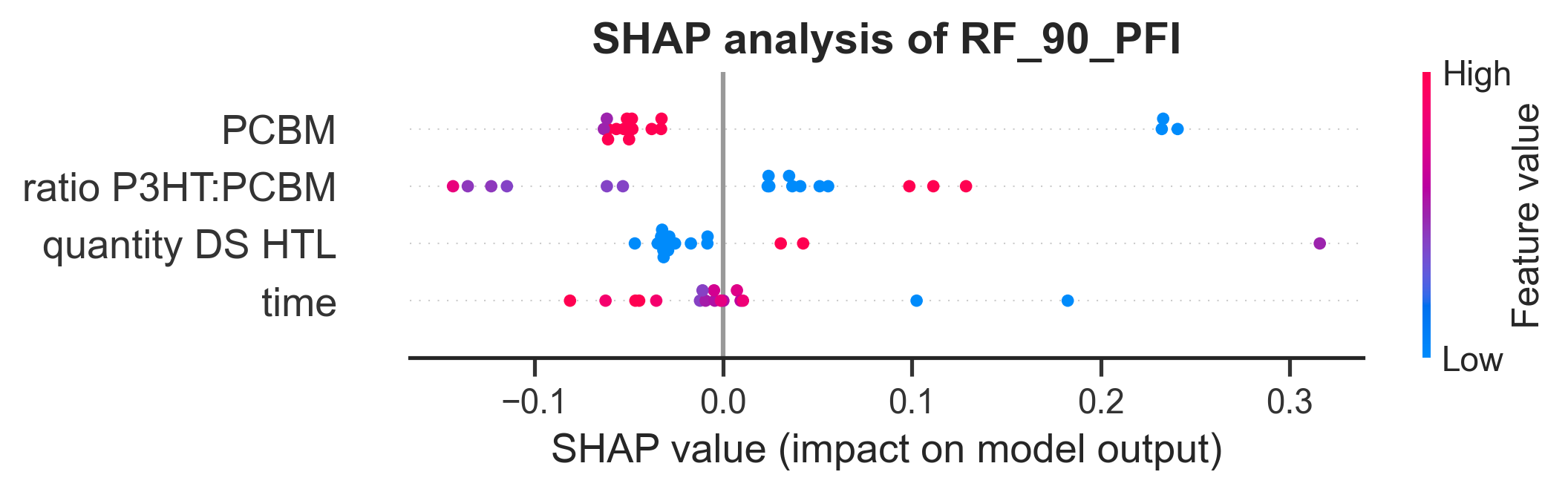}\label{fig:shap4}}    
                 {   \caption{SHAP results for the best {ML} model generated with data up to 180 days.} \label{fig:SHAP}}
                    
                   
\end{figure}

\section{Discussion and conclusions}\label{sec:conclusions}

This paper has presented the application of optimal ML framework{s} to characterize the degradation behavior, in terms of PCE, of OSCs with multilayer structure ITO/PEDOT:PSS/P3HT:PCBM/Al. A dataset {with {166 entries} entries was created, containing PCE values of various OSC devices measured over 180 days. This dataset was supplemented with seven variables describing environmental conditions of the experiments and manufacturing parameters of the devices.} Through hyper-optimization of a set of ML models we have presented an accuracy analysis of different methods, which were fed with OSC data periodized into sets from 30 to 180 days. The benchmarking has confirmed the validity of models like RF or GB to confer R$^2$ values over 0.90, reaching in some extents R$^2$$\sim$0.96-0.97 {and} error metrics (RMSE, SSE and MAE) {significantly low} when long term data is used for training. To reinforce the suitability of these ML models, classical {LS} regression fitting methods have been {compared. These proved not to be suitable for a multivariable dataset like ours, especially when dealing with long term data of the OSCs. Consequently, their ability to predict PCE values is highly unreliable.} 

ML models proved to offer high feasibility to predict the behavior of unknown OSCs. {Feature analysis suggests that the most influential variable is} the solvent in the HTL layer, i.e., the amount of PEDOT:PSS. This is explained as the multilayer structure of the OSCs needs a minimum value of PEDOT:PSS to ensure a layer that completely covers the substrate. Moreover, the ratio P3HT:PCBM also exhibits significant importance, being higher values representative of greater impact on the model. Finally, it has been {observed} that variables such as temperature, humidity, dew point and pressure have lesser impact on the models, explained by the encapsulation made to the OSCs during their manufacturing.

\section*{CRediT authorship contribution statement}
{
{Conceptualization D.V. and J.V.A.-R.; Methodology D.V., J.V.A.-R., D.D. and F.R.-M.; Software D.V., J.V.A.-R., D.D. and M.F.; Data curation D.V. and M.F.; Formal analysis D.V., F.R.-M.; Funding acquisition D.V., J.C.F. and J.V.A.-R.; Investigation D.V., F.R.-M. and J.V.A.-R.; Resources F.R.-M. and J.C.F.; Supervision J.C.F. and J.V.A.-R.; Validation D.D. and M.F.; Visualization D.V. and M.F; Writing - Original draft D.V. and J.V.A.-R.; Writing - Review \& Editing J.V.A.-R., D.D. and F.R.-M.}
}
\section*{Acknowledgements}

{This work is part of the project CIAICO/2023/193 funded by Generalitat Valenciana. The authors acknowledge that support.}

J.V.A.-R. and D.D. acknowledge Gobierno de Arag\'on Fondo Social Europeo (Research Group E07\_23R), the State Research Agency of Spain (MCIN/ AEI/ 10.13039/501100011033/ FEDER, UE) for financial support (PID2022-140159NA-I00) and the European Union's Recovery and Resilience Facility-Next Generation in the framework of the General Invitation of the Spanish Government's public business entity Red.es to participate in talent attraction and retention programmes within Investment 4 of Component 19 of the Recovery, Transformation and Resilience Plan (MOMENTUM, MMT24-ISQCH-01).

\appendix
{\section{{Extended accuracy metrics of ML models}}\label{sec:app}
{
This section comprises extended accuracy metrics of the ML models used in this work, when the temporal scope of the training-validation data is varied: 30, 60, 90, 120 and 150 days, respectively. 

            \begin{table}[h!]
                    \caption{Accuracy metrics of the ML regression models computed with training data up to 30 days.}
                    \label{tab:table4}
                    \centering
                    \footnotesize
                    \begin{tabular}{r|rcccc}
                    \toprule
                    \textbf{Training data}&\textbf{ML model}&\textbf{R$^2$}&\textbf{RMSE}&\textbf{SSE}&\textbf{MAE}\\
                    \midrule
                    30 days
                    		&GB-80-20&0.86&0.140&0.0647&0.1200\\ 
                   		&GB-70-30&0.73&0.140&0.0622&0.1200\\ 
            		&GB-60-40&0.84&0.160&0.1103&0.1500\\ 
            		\cline{2-6} 
                   		&NN-80-20&0.76&0.180&0.1655&0.1800\\ 
            		&NN-70-30&0.51&0.170&0.0832&0.1600\\ 
            		&NN-60-40&0.62&0.170&0.0931&0.1400\\ 
            		\cline{2-6} 
            		&MVL-80-20&0.71&0.200&0.0984&0.1500\\ 
            		&MVL-70-30&0.62&0.220&0.1006&0.1400\\ 
            		&MVL-60-40&0.66&0.170&0.1067&0.1400\\ 
            		\cline{2-6} 
            		&RF-80-20&0.86&0.130&0.0573&0.1000\\ 
                   		&RF-70-30&0.78&0.130&\textbf{0.0558}&0.1000\\ 
                   		&\textbf{RF-60-40}&\textbf{0.89}&\textbf{0.120}&0.0725&\textbf{0.0950}\\    
                    \bottomrule
                    \end{tabular}
                    \end{table}
            
            \begin{table}[h!]
                    \caption{Accuracy metrics of the ML regression models computed with training data up to 60 days.}
                    \label{tab:table5}
                    \centering
                    \footnotesize
                    \begin{tabular}{r|rcccc}
                    \toprule
                    \textbf{Training data}&\textbf{ML model}&\textbf{R$^2$}&\textbf{RMSE}&\textbf{SSE}&\textbf{MAE}\\
                    \midrule
                    60 days
                    		&GB-90-10&0.85&0.1000&\textbf{0.0155}&\textbf{0.0800}\\
            &GB-80-20&0.76& 0.1500&0.0433&0.1300\\
            &GB-70-30&0.84&0.1500&0.0597&0.1300\\
            &GB-60-40&0.78&0.1700&0.1275&0.1500\\
            \cline{2-6} 
            &NN-90-10&0.89&0.1300&0.1239&	0.1000\\
            &NN-80-20&0.86&0.1700	&0.1677&	0.1500\\
            &NN-70-30&0.88&0.1500	&0.0506&	0.1300\\
            &NN-60-40&0.79&0.1600	&0.0677&	0.1400\\
            \cline{2-6} 
            &MVL-90-10&0.71&0.1500&0.1059&0.1300\\
            &MVL-80-20&0.80&0.1700&0.1120&0.1500\\
            &MVL-70-30&0.63&0.1700&0.1117&0.1500\\
            &MVL-60-40&0.61&0.1900&0.1290&0.1500\\
            \cline{2-6} 
            &\textbf{RF-90-10}&\textbf{0.98}&\textbf{0.1000}&0.0702&0.0850\\
            &RF-80-20&0.92&0.1400&	0.0832&0.1300\\
            &RF-70-30&0.78&0.1600&	0.0747&0.1400\\
            &RF-60-40&0.81&0.1300&	0.0578&0.1000\\
                    \bottomrule
                    \end{tabular}
                    \end{table}
            
                    \begin{table}[h!]
                    \caption{Accuracy metrics of the ML regression models computed with training data up to 90 days.}
                    \label{tab:table6}
                    \centering
                    \footnotesize
                    \begin{tabular}{r|rrccc}
                    \toprule
                    \textbf{Training data}&\textbf{ML model}&\textbf{R$^2$}&\textbf{RMSE}&\textbf{SSE}&\textbf{MAE}\\
                    \midrule
                    90 days
                    		&GB-90-10&0.87&\textbf{0.0730}&\textbf{0.0255}&\textbf{0.0660}\\
            		&GB-80-20&0.87&0.1100&0.0736&0.0860\\
                   		&GB-70-30&0.88&0.0940&0.0775&0.0770\\
            		&GB-60-40&0.75&0.1400&0.0755&0.1100\\
            		\cline{2-6} 
                   		&NN-90-10&0.86&0.0790&0.0332&0.0630\\
            		&NN-80-20&0.82&0.1300&0.1595&0.1200\\
            		&NN-70-30&0.46&0.1900&0.1071&0.1700\\
            		&NN-60-40&0.52&0.1900&0.1611&0.1500\\
            		\cline{2-6} 
            		&MVL-90-10&0.61&0.1100&0.3161&0.1000\\
            		&MVL-80-20&0.63&0.1700&0.1468&0.1400\\
            		&MVL-70-30&0.54&0.1700&0.1378&0.1400\\
            		&MVL-60-40&0.55&0.1800&0.1569&0.1500\\
            		\cline{2-6} 
            		&\textbf{RF-90-10}&\textbf{0.89}&0.0750&0.0300&0.0670\\
            		&RF-80-20&0.87&0.0940&0.0909&0.0700\\
                   		&RF-70-30&0.80&0.1000&0.0613&0.0780\\
                   		&RF-60-40&0.74&0.1400&0.0860&0.1100\\
                    \bottomrule
                    \end{tabular}
                    \end{table}	
            
            \begin{table}[h!]
                    \caption{Accuracy metrics of the ML models computed with training data up to 120 days.}
                    \label{tab:table7}
                    \centering
                    \footnotesize
                    \begin{tabular}{r|rcccc}
                    \toprule
                    \textbf{Training data}&\textbf{ML model}&\textbf{R$^2$}&\textbf{RMSE}&\textbf{SSE}&\textbf{MAE}\\
                    \midrule
                    120 days
                    		&GB-90-10&0.95&0.0670&\textbf{0.0312}&	0.0540\\
            		&GB-80-20&0.88&0.0700&	0.0324&	0.0510\\
                   		&GB-70-30&0.89&0.0810&	0.0453&	0.0620\\
            		&GB-60-40&0.81&0.1300&	0.1447&	0.1100\\
            		\cline{2-6} 
                   		&NN-90-10&0.88&0.0600&	0.2140&	0.0450\\
            		&NN-80-20&0.83&0.0890&	0.0576&	0.0660\\
            		&NN-70-30&0.71&0.1300&	0.0767&	0.0960\\
            		&NN-60-40&0.59&0.1700&	0.1743&	0.1300\\
            		\cline{2-6} 
            		&MVL-90-10&0.70&0.0990&	0.1729&	0.0850\\
            		&MVL-80-20&0.65&0.1400&	0.1820&	0.1100\\
            		&MVL-70-30&0.60&0.1600&	0.2311&	0.1400\\
            		&MVL-60-40&0.51&0.1800&	0.2370&	0.1500\\
            		\cline{2-6} 
            		&\textbf{RF-90-10}&\textbf{0.98}&\textbf{0.0300}&	0.0507&	\textbf{0.0240}\\
            		&RF-80-20&0.88&0.0680&	0.0651&	0.0480\\
                   		&RF-70-30&0.86&0.0920&	0.0609&	0.0690\\
                   		&RF-60-40&0.84&0.1000&	0.0574&	0.0770\\
                    \bottomrule
                    \end{tabular}
                    \end{table}
            
            \begin{table}[h!]
                    \caption{Accuracy metrics of the ML regression models computed with training data up to 150 days.}
                    \label{tab:table8}
                    \centering
                    \footnotesize
             \begin{tabular}{r|rcccc}
                    \toprule
                    \textbf{Training data}&\textbf{ML model}&\textbf{R$^2$}&\textbf{RMSE}&\textbf{SSE}&\textbf{MAE}\\
                    \midrule
                    150 days
                    		&GB-90-10&0.98&0.0300&	0.0510&	0.0230\\
            		&GB-80-20&0.93&0.0660&	0.0242&	0.0420\\
                   		&GB-70-30&0.94&0.0730&	0.0351&	0.0580\\
            		&GB-60-40&0.91&0.0820&	0.0591&	0.0640\\
            		\cline{2-6} 
                   		&NN-90-10&0.90&0.0580&	0.0760&	0.0540\\
            		&NN-80-20&0.82&0.0110&	0.1277&	0.0630\\
            		&NN-70-30&0.88&0.0960&	0.0871&	0.0740\\
            		&NN-60-40&0.78&0.1200&	0.1533&	0.0910\\
            		\cline{2-6} 
            		&MVL-90-10&0.84&0.0920&	0.2113&	0.0800\\
            		&MVL-80-20&0.87&0.1100&	0.2139&	0.0840\\
            		&MVL-70-30&0.88&0.1200&	0.2163&	0.0900\\
            		&MVL-60-40&0.82&0.1100&	0.2184&	0.0910\\
            		\cline{2-6} 
            		&\textbf{RF-90-10}&\textbf{0.97}&\textbf{0.0290}&	\textbf{0.0219}&	\textbf{0.0210}\\
            		&RF-80-20&0.96&0.0510&	0.0394&	0.0324\\
                   		&RF-70-30&0.94&0.0780&	0.0417&	0.0570\\
                   		&RF-60-40&0.90&0.0930&	0.1069&	0.0943\\
                    \bottomrule
                    \end{tabular}
                    \end{table}

}

\clearpage
\section{OSC database}\label{sec:app2}
{Database in $csv$ file consisting of {166} entries which includes up to seven variables regarding both the manufacturing process and environmental conditions for more than 180 days (Table~\ref{tab:table1}). PCE values of several polymeric OSCs with a multilayer structure ITO/PEDOT:PSS/P3HT:PCBM/Al were measured.}

{In addition, to ensure reproducibility, another $csv$ file provides a dataset without data of $Cell4$. Finally, another $csv$ file can be used to test the predictive ability of the model to predict data of such cell (see Section~\ref{sec:preML}.)}

\section{ROBERT report}\label{sec:app3}
{{
Detailed report obtained with ROBERT (v1.0.6) after modeling the entire database. Results and reproducibility details are contained in this file.
}
%
%
}

\section*{Data availability}
{
{The program ROBERT (v1.0.6) to model the database is publicly available at:} {\url{https://github.com/jvalegre/robert/releases}}
 
{Its online documentation~\cite{Readthedocs} at:} {\url{https://robert.readthedocs.io/en/latest/index.html}}

{The datasets are available in~\ref{sec:app2}.}
}

             \bibliographystyle{elsarticle-num}
                         \biboptions{numbers}

{\small
\bibliography{2025_ESWA}}

            \end{document}